\pdfoutput=1
\documentclass[10pt,twocolumn,letterpaper]{article}

\usepackage{cvpr}
\usepackage{times}
\usepackage{epsfig}
\usepackage{graphicx}
\usepackage{amsmath}
\usepackage{amssymb}
\usepackage{multirow}
\usepackage{makecell}
\usepackage{url}
\usepackage{booktabs}
\usepackage[table]{xcolor}
\usepackage{caption}
\usepackage{subcaption}

\cvprfinalcopy 

\def\cvprPaperID{2401} 

\ifcvprfinal\fi
\begin{document}

\title{Deep Modular Co-Attention Networks for Visual Question Answering}

\author{
Zhou Yu\textsuperscript{\rm 1}\quad\quad
Jun Yu\textsuperscript{\rm 1}\thanks{Jun Yu is the corresponding author}\quad\quad
Yuhao Cui\textsuperscript{\rm 1}\quad\quad
Dacheng Tao\textsuperscript{\rm 2}\quad\quad
Qi Tian\textsuperscript{\rm 3}
\\
\normalsize\textsuperscript{\rm 1}Key Laboratory of Complex Systems Modeling and Simulation, \\
\normalsize School of Computer Science and Technology, Hangzhou Dianzi University, China.\\
\normalsize \textsuperscript{\rm 2}UBTECH Sydney AI Centre, School of Computer Science, FEIT, University of Sydney, Australia\\
\normalsize \textsuperscript{\rm 3}Noah's Ark Lab, Huawei, China\\
\tt\small \{yuz, yujun, cuiyh\}@hdu.edu.cn,~dacheng.tao@sydney.edu.au,~tian.qi1@huawei.com
}

\maketitle
\thispagestyle{empty}

\begin{abstract}
Visual Question Answering (VQA) requires a fine-grained and simultaneous understanding of both the visual content of images and the textual content of questions. Therefore, designing an effective `co-attention' model to associate key words in questions with key objects in images is central to VQA performance. So far, most successful attempts at co-attention learning have been achieved by using shallow models, and deep co-attention models show little improvement over their shallow counterparts. In this paper, we propose a deep Modular Co-Attention Network (MCAN) that consists of Modular Co-Attention (MCA) layers cascaded in depth. Each MCA layer models the self-attention of questions and images, as well as the guided-attention of images jointly using a modular composition of two basic attention units. We quantitatively and qualitatively evaluate MCAN on the benchmark VQA-v2 dataset and conduct extensive ablation studies to explore the reasons behind MCAN's effectiveness. Experimental results demonstrate that MCAN significantly outperforms the previous state-of-the-art. Our best single model delivers 70.63$\%$ overall accuracy on the test-dev set. Code is available at \url{https://github.com/MILVLG/mcan-vqa}.
\end{abstract}

\section{Introduction}
Multimodal learning to bridge vision and language has gained broad interest from both the computer vision and natural language processing communities. Significant progress has been made in many vision-language tasks, including image-text matching \cite{nam2016dual,kim2018bilinear}, visual captioning \cite{donahue2015long,xu2015show,anderson2017up-down}, visual grounding \cite{fukui2016multimodal,yu2018rethinking} and visual question answering (VQA) \cite{antol2015vqa,malinowski2014multi,kim2018bilinear,zhao2018open}. Compared to other multimodal learning tasks, VQA is a more challenging task that requires fine-grained semantic understanding of both the image and the question, together with visual reasoning to predict an accurate answer.

\begin{figure}
\begin{center}
\includegraphics[width=0.4\textwidth]{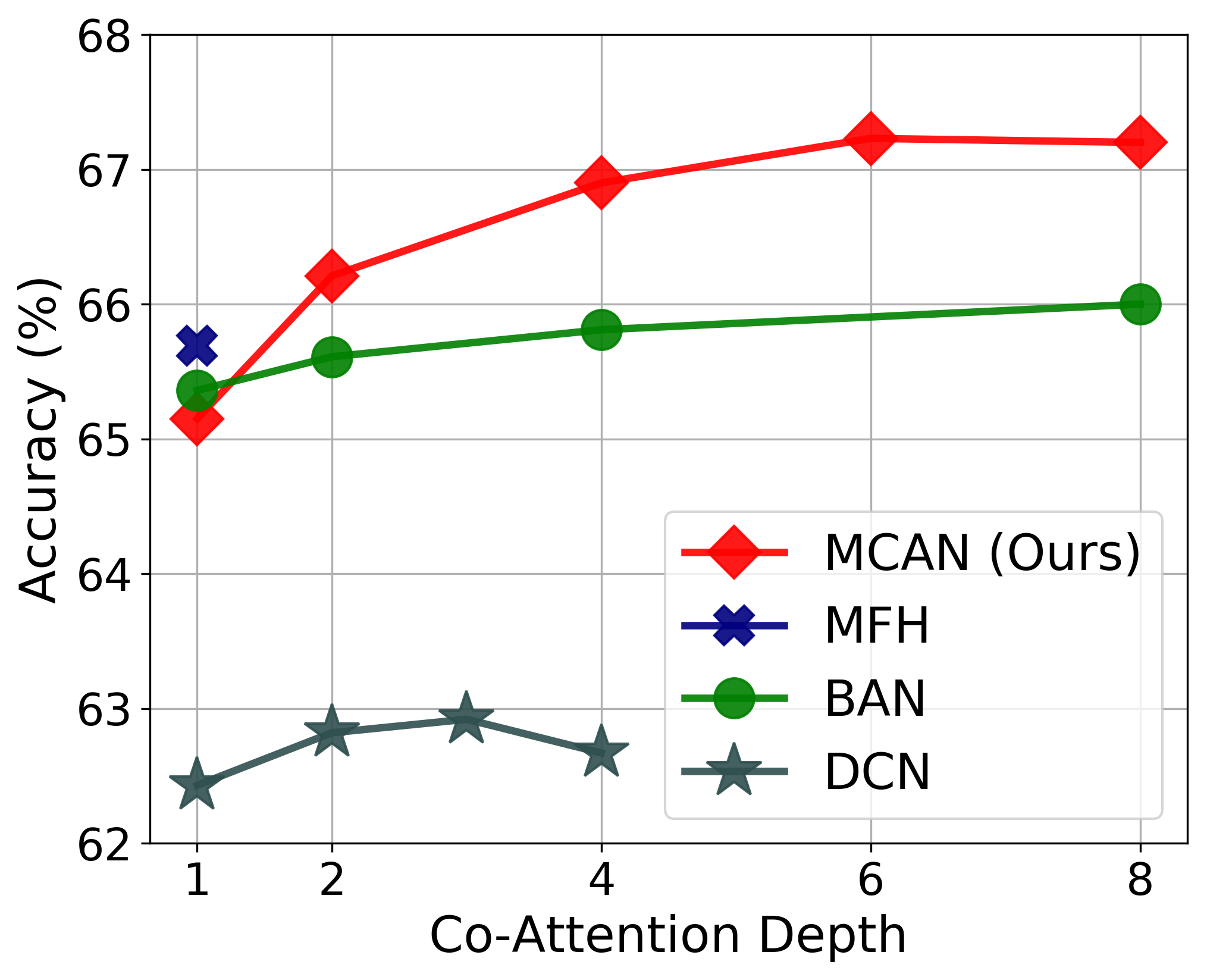}
\vspace{-5pt}
\caption{Accuracies \textit{vs}. co-attention depth on VQA-v2 \emph{val} split. We list most of the state-of-the-art approaches with (deep) co-attention models. Except for DCN \cite{nguyen2018improved} which uses the convolutional visual features and thus leads to inferior performance, all the compared methods (\emph{i.e.}, MCAN, BAN \cite{kim2018bilinear} and MFH \cite{yu2018beyond}) use the same bottom-up attention visual features to represent images \cite{anderson2017up-down}.}
\vspace{-25pt}
\label{fig:depth_acc}
\end{center}
\end{figure}

The attention mechanism is a recent advance in deep neural networks, that has successfully been applied to the unimodal tasks (\eg, vision \cite{mnih2014recurrent}, language \cite{bahdanau2014neural}, and speech \cite{chorowski2015attention}),
as well as the aforementioned multimodal tasks.
The idea of learning visual attention on image regions from the input question in VQA was first proposed by \cite{shih2016look,chen2015abc}, and it becomes a {de-facto} component of almost all VQA approaches \cite{fukui2016multimodal,kim2016hadamard,anderson2017up-down}. Along with visual attention, learning textual attention on the question key words is also very important. Recent works have shown that simultaneously learning co-attention for the visual and textual modalities can benefit the fine-grained representation of the image and question, leading to more accurate prediction \cite{lu2016hierarchical,yu2018beyond}. However, these co-attention models learn the coarse interactions of multimodal instances, and the learned co-attention cannot infer the correlation between each image region and each question word. This results in a significant limitation of these co-attention models.

To overcome the problem of insufficient multimodal interactions, two dense co-attention models BAN \cite{kim2018bilinear} and DCN \cite{nguyen2018improved} have been proposed to model dense interactions between any image region and any question word. The dense co-attention mechanism facilitates the understanding of image-question relationship to correctly answer questions. Interestingly, both of these dense co-attention models can be cascaded in depth, form deep co-attention models that support more complex visual reasoning, thereby potentially improving VQA performance. However, these deep models shows little improvement over their corresponding shallow counterparts or the coarse co-attention model MFH \cite{yu2018beyond} (see Figure \ref{fig:depth_acc}). We think the bottleneck in these deep co-attention models is a deficiency of simultaneously modeling dense self-attention within each modality (\emph{i.e.}, word-to-word relationship for questions, and region-to-region relationship for images).

Inspired by the Transformer model in machine translation \cite{vaswani2017attention}, here we design two general attention units: a self-attention (SA) unit that can model the dense intra-modal interactions (word-to-word or region-to-region); and a guided-attention (GA) unit to model the dense inter-modal interactions (word-to-region). After that, by modular composition of the SA and GA units, we obtain different Modular Co-Attention (MCA) layers that can be cascaded in depth. Finally, we propose a deep Modular Co-Attention Network (MCAN) which consists of cascaded MCA layers. Results in Figure \ref{fig:depth_acc} shows that a deep MCAN model significantly outperforms existing state-of-the-art co-attention models on the benchmark VQA-v2 dataset \cite{goyal2016making}, which verifies the synergy of self-attention and guided-attention in co-attention learning, and also highlights the potential of deep reasoning. Furthermore, we find that modeling self-attention for image regions can greatly improve the object counting performance, which is challenging for VQA.

\section{Related Work}
We briefly review previous studies on VQA, especially those studies that introduce co-attention models.
\\
\textbf{Visual Question Answering (VQA).} VQA has been of increasing interest over the last few years.
The multimodal fusion of global features are the most straightforward VQA solutions. The image and question are first represented as global features and then fused by a multimodal fusion model to predict the answer \cite{zhou2015simple}. Some approaches introduce a more complex model to learn better question representations with LSTM networks \cite{antol2015vqa}, or a better multimodal fusion model with residual networks \cite{kim2016multimodal}.

One limitation of the aforementioned multimodal fusion models is that the global feature representation of an image may lose critical information to correctly answer the questions about local image regions (\eg, ``what is in the woman's left hand''). Therefore, recent approaches have introduced the {visual attention} mechanism into VQA by adaptively learning the attended image features for a given question, and then performing multimodal feature fusion to obtain the accurate prediction. Chen \etal proposed a question-guided attention map that projected the question embeddings into the visual space and formulated a configurable convolutional kernel to search the image attention region \cite{chen2015abc}. Yang \etal proposed a stacked attention network to learn the attention iteratively \cite{yang2016stacked}. Fukui \etal \cite{fukui2016multimodal}, Kim \etal \cite{kim2016hadamard}, Yu \etal \cite{yu2017mfb, yu2018beyond} and Ben \etal \cite{ben2017mutan} exploited different multimodal bilinear pooling methods to integrate the visual features from the image's spatial grids with the textual features from the questions to predict the attention. Anderson \etal introduced a bottom-up and top-down attention mechanism to learn the attention on candidate objects rather than spatial grids \cite{anderson2017up-down}.
\\
\textbf{Co-Attention Models.} Beyond understanding the visual contents of the image, VQA also requires to fully understand the semantics of the natural language question. Therefore, it is necessary to learn the {textual attention} for the question and the visual attention for the image simultaneously. Lu \etal proposed a co-attention learning framework to alternately learn the image attention and question attention \cite{lu2016hierarchical}. Yu \etal reduced the co-attention method into two steps, self-attention for a question embedding and the question-conditioned attention for a visual embedding \cite{yu2018beyond}.  Nam \etal proposed a multi-stage co-attention learning model to refine the attentions based on memory of previous attentions \cite{nam2016dual}. However, these co-attention models learn separate attention distributions for each modality (image or question), and neglect the dense interaction between each question word and each image region. This become a bottleneck for understanding fine-grained relationships of multimodal features. To address this issue, dense co-attention models have been proposed, which establish the complete interaction between each question word and each image region \cite{nguyen2018improved,kim2018bilinear}. Compared to the previous co-attention models with coarse interactions, the dense co-attention models deliver significantly better VQA performance.

\section{Modular Co-Attention Layer}
Before presenting the Modular Co-Attention Network, we first introduce its basic component, the Modular Co-Attention (MCA) layer. The MCA layer is a modular composition of the two basic attention units, \ie, the self-attention (SA) unit and the guided-attention (GA) unit, inspired by the \emph{scaled dot-product attention} proposed in \cite{vaswani2017attention}. Using different combinations, we obtain three MCA variants with different motivations.

\subsection{Self-Attention and Guided-Attention Units}
The input of scaled dot-product attention consists of queries and keys of dimension $d_{\mathrm{key}}$, and values of dimension $d_{\mathrm{value}}$. For simplicity, $d_{\mathrm{key}}$ and $d_{\mathrm{value}}$ are usually set to the same number $d$. We calculate the dot products of the query with all keys, divide each by $\sqrt{d}$ and apply a softmax function to obtain the attention weights on the values. Given a query $q\in\mathbb{R}^{1\times d}$, $n$ key-value pairs (packed into a key matrix $K\in\mathbb{R}^{n\times d}$ and a value matrix $V\in\mathbb{R}^{n\times d}$), the attended feature $f\in\mathbb{R}^{1 \times d}$ is obtained by weighted summation over all values $V$ with respect to the attention learned from $q$ and $K$:
\begin{equation}\label{eq:scaled_dot_product}
f = A(q,K,V)=\mathrm{softmax}(\frac{qK^T}{\sqrt{d}})V
\end{equation}

\captionsetup[subfigure]{font=small}
\begin{figure}
    \centering
    \begin{subfigure}[h]{0.4\columnwidth}
        \includegraphics[width=\linewidth]{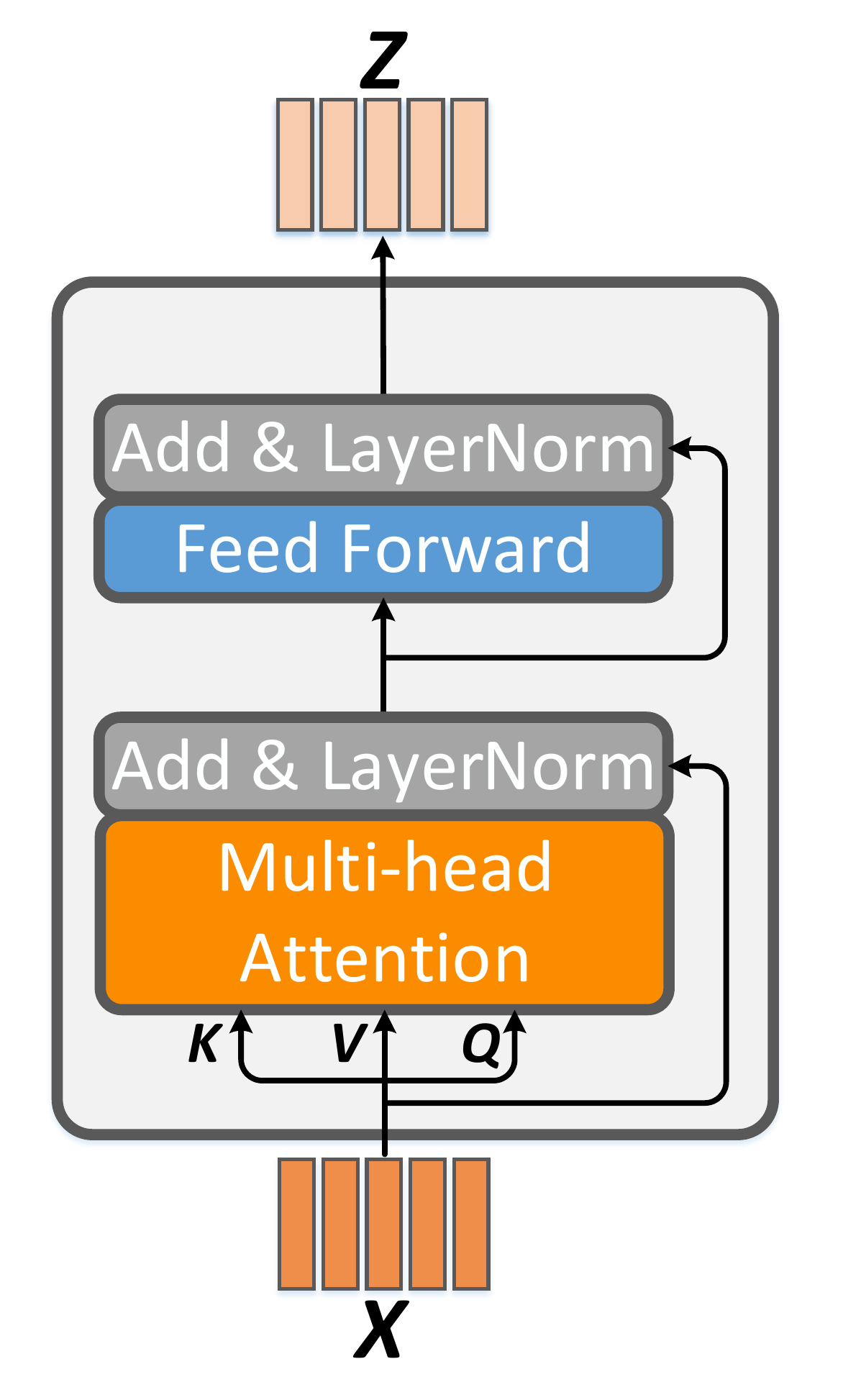}
        \caption{Self-Attention (SA)}\label{fig:sa}
    \end{subfigure}
    \begin{subfigure}[h]{0.42\columnwidth}
        \includegraphics[width=\linewidth]{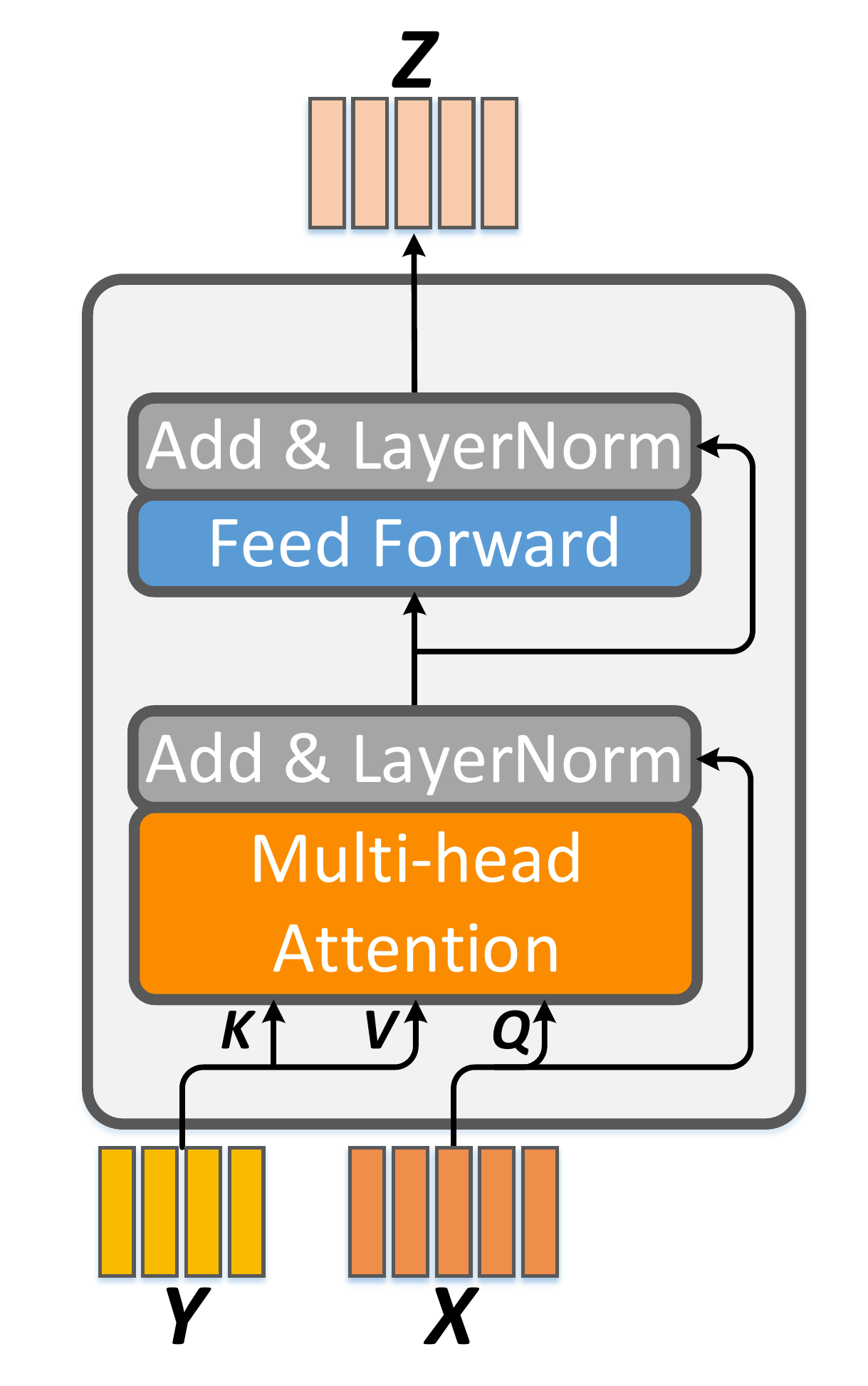}
        \caption{Guided-Attention (GA)}\label{fig:ga}
    \end{subfigure}
    \vspace{-5pt}
    \caption{Two basic attention units with multi-head attention for different types of inputs. SA takes one group of input features $X$ and output the attended features $Z$ for $X$; GA takes two groups of input features $X$ and $Y$ and output the attended features $Z$ for $X$ guided by $Y$.}
    \vspace{-10pt}
    \label{fig:att_unit}
\end{figure}

To further improve the representation capacity of the attended features, \emph{multi-head attention} is introduced in \cite{vaswani2017attention}, which consists of $h$ paralleled `heads'. Each head corresponds to an independent scaled dot-product attention function. The attended output features $f$ is given by:
\begin{equation}\label{eq:ma}
f = MA(q,K,V)=[\mathrm{head}_1,\mathrm{head}_2,...,\mathrm{head}_h]W^o
\end{equation}

\begin{equation}\label{eq:ma_head}
\mathrm{head}_j=A(qW_j^Q,KW_j^K,VW_j^V)
\end{equation}
where $W_j^Q, W_j^K, W_j^V \in\mathbb{R}^{d \times d_h}$ are the projection matrices for the $j$-th head, and $W_o\in\mathbb{R}^{h*d_h \times d}$. $d_h$ is the dimensionality of the output features from each head. To prevent the multi-head attention model from becoming too large, we usually have $d_h=d/h$.
In practice, we can compute the attention function on a set of $m$ queries $Q=[q_1;q_2;...;q_m]\in\mathbb{R}^{m \times d}$ seamlessly by replacing $q$ with $Q$ in Eq.(\ref{eq:ma}), to obtain the attended output features $F\in\mathbb{R}^{m\times d}$.

We build two attention units on top of the multi-head attention to handle the multimodal input features for VQA, namely the \emph{self-attention} (SA) unit and the \emph{guided-attention} (GA) unit. The SA unit (see Figure \ref{fig:sa}) is composed of a multi-head attention layer and a pointwise feed-forward layer. Taking one group of input features $X=[x_1;...;x_m]\in\mathbb{R}^{m \times d_x}$, the multi-head attention learns the pairwise relationship between the paired sample $<x_i,x_j>$ within $X$ and outputs the attended output features $Z\in\mathbb{R}^{m\times d}$ by weighted summation of all the instances in $X$. The feed-forward layer takes the output features of the multi-head attention layer, and further transforms them through two fully-connected layers with ReLU activation and dropout (FC($4d$)-ReLU-Dropout(0.1)-FC($d$)). Moreover, residual connection \cite{he2015deep} followed by layer normalization \cite{ba2016layer} is applied to the outputs of the two layers to facilitate optimization.
The GA unit (see Figure \ref{fig:ga}) takes two groups of input features $X\in\mathbb{R}^{m \times d_x}$ and $Y=[y_1;...;y_n]\in\mathbb{R}^{n \times d_y}$, where $Y$ guides the attention learning for $X$. Note that the shapes of $X$ and $Y$ are flexible, so they can be used to represent the features for different modalities (\eg, questions and images). The GA unit models the pairwise relationship between the each paired sample $<x_i,y_j>$ from $X$ and $Y$, respectively.
\captionsetup[subfigure]{font=scriptsize}
\begin{figure}
    \centering
    \begin{subfigure}[h]{0.325\columnwidth}
        \includegraphics[width=\linewidth]{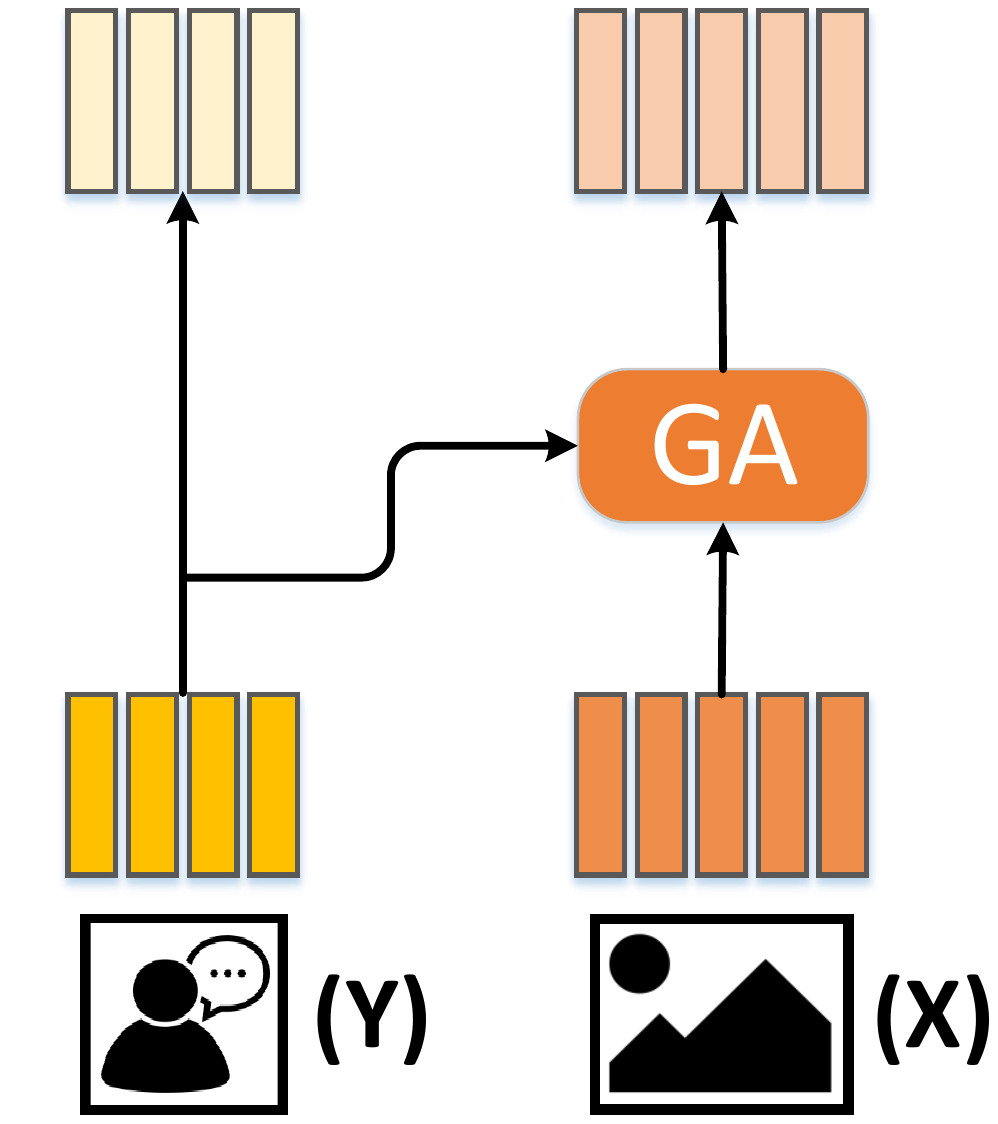}
        \caption{ID(Y)-GA(X,Y)}\label{fig:cau_ga}
    \end{subfigure}
    \begin{subfigure}[h]{0.325\columnwidth}
        \includegraphics[width=\linewidth]{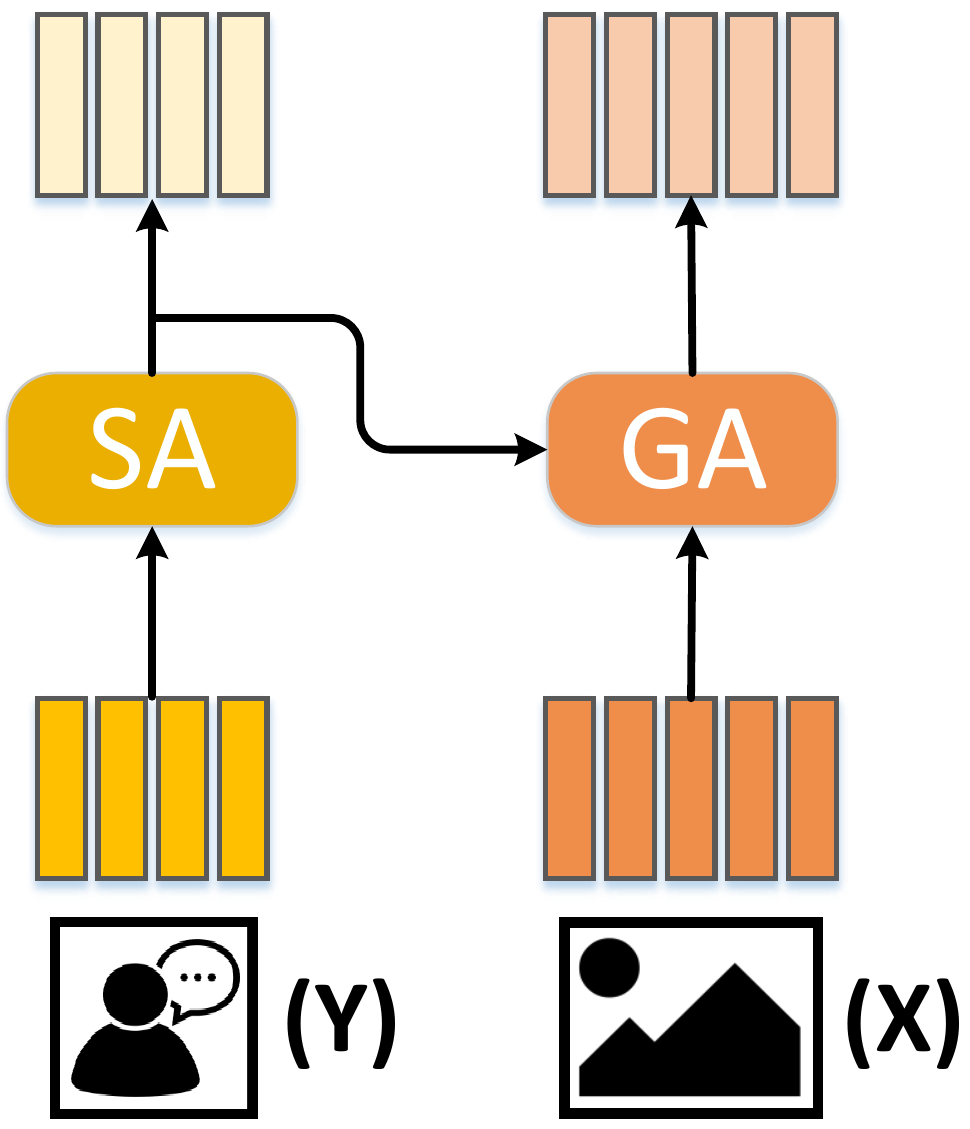}
        \caption{SA(Y)-GA(X,Y)}\label{fig:cau_sa_ga}
    \end{subfigure}
    \begin{subfigure}[h]{0.325\columnwidth}
        \includegraphics[width=\linewidth]{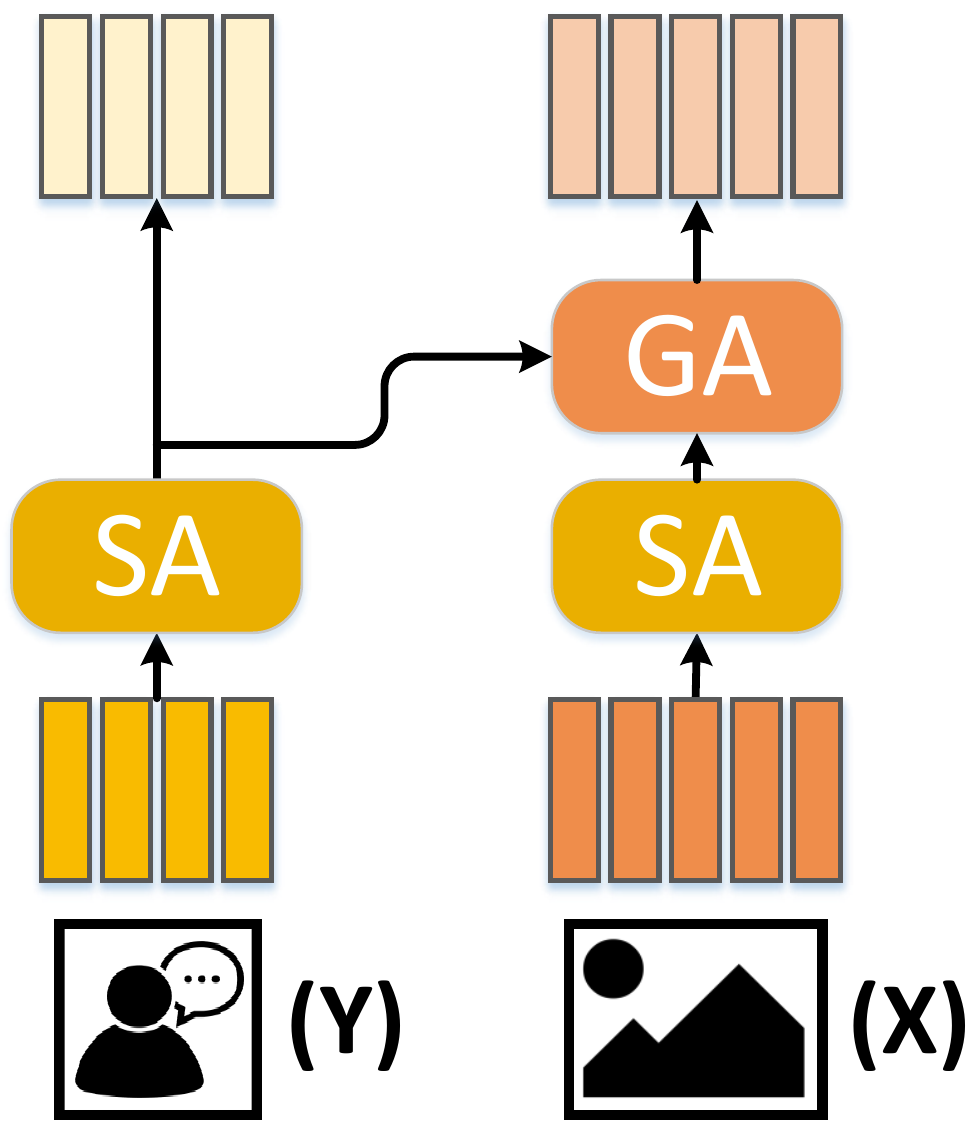}
        \caption{SA(Y)-SGA(X,Y)}\label{fig:cau_sa_sga}
    \end{subfigure}
    \vspace{-5pt}
    \caption{Flowcharts of three MCA variants for VQA. (Y) and (X) denote the question and image features respectively.}
    \vspace{-10pt}
    \label{fig:cau_arch}
\end{figure}

\begin{figure*}
\begin{center}
\includegraphics[width=0.95\textwidth]{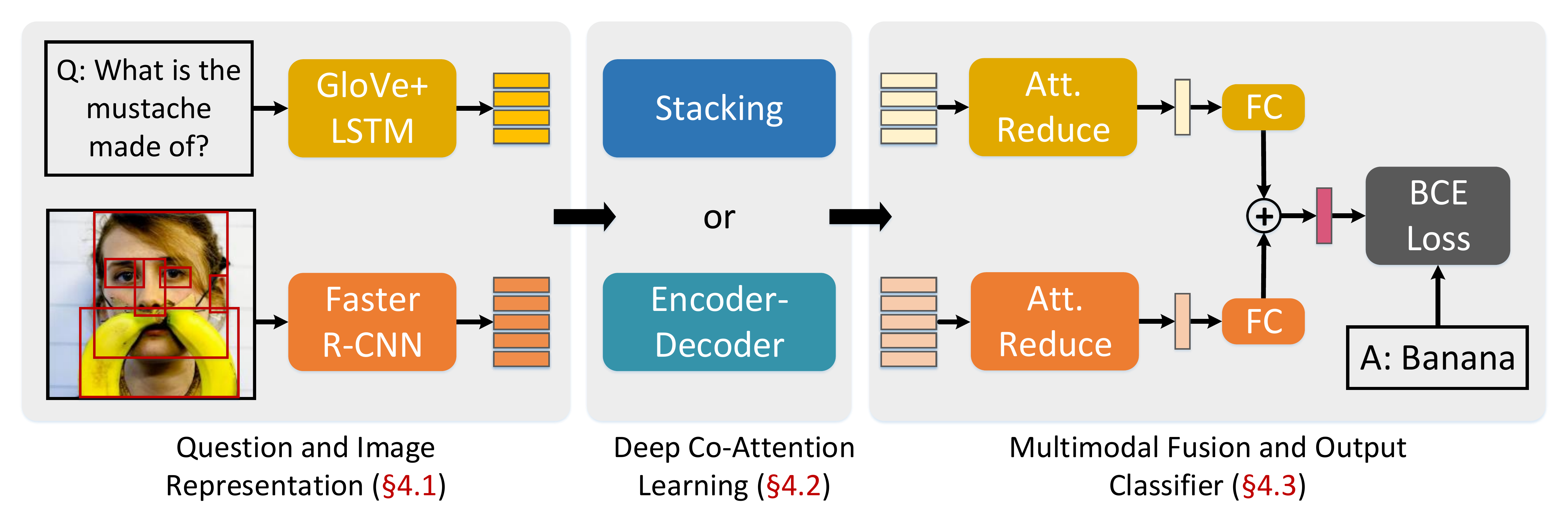}
\caption{Overall flowchart of the deep Modular Co-Attention Networks (MCAN). In the Deep Co-attention Learning stage, we have two alternative strategies for deep co-attention learning, namely \emph{stacking} and \emph{encoder-decoder}.}
\vspace{-10pt}
\label{fig:mcan_flowchart}
\end{center}
\vspace{-10pt}
\end{figure*}

\noindent\textbf{Interpretation:} Since the multi-head attention in Eq.(\ref{eq:ma}) plays a key role in the two attention units, we take a closer look at it to see how it works with respect to different types of inputs. For a SA unit with input features $X$, for each $x_i\in X$, its attended feature $f_i=\mathrm{MA}(x_i,X,X)$ can be understood as \emph{reconstructing} $x_i$ by all the samples in $X$ with respect to their normalized similarities to $x_i$.
Analogously, for a GA unit with input features $X$ and $Y$, the attended feature $f_i=\mathrm{MA}(x_i,Y,Y)$ for $x_i\in X$ is obtained by reconstructing $x_i$ by all the samples in $Y$ with respect to their normalized cross-modal similarity to $x_i$.

\subsection{Modular Composition for VQA}
Based on the two basic attention units in Figure \ref{fig:att_unit}, we composite them to obtain three modular co-attention (MCA) layers (see Figure \ref{fig:cau_arch}) to handle the multimodal features for VQA. All three MCA layers can be cascaded in depth, such that the outputs of the previous MCA layer can be directly fed to the next MCA layer. This implies that the number of input features is equal to the number of output features without instance reduction.

The ID(Y)-GA(X,Y) layer in Figure \ref{fig:cau_ga} is our baseline. In ID(Y)-GA(X,Y), the input question features are directly passed through to the output features with an identity mapping, and the dense inter-modal interaction between each region $x_i\in X$ with each word $y_i\in Y$ is modeled in a GA(X,Y) unit. These interactions are further exploited to obtain the attended image features. Compared to the ID(Y)-GA(X,Y) layer, the SA(Y)-GA(X,Y) layer in Figure \ref{fig:cau_sa_ga} adds a SA(Y) unit to model the dense intra-modal interaction between each question word pair $\{y_i,y_j\}\in Y$. The SA(Y)-SGA(X,Y) layer in Figure \ref{fig:cau_sa_sga} continues to add a SA(X) unit to the SA(Y)-GA(X,Y) layer to model the intra-modal interaction between each image region pairs $\{x_i,x_j\}\in X$.\footnote{In our implementation, we omit the feed-forward layer and norm layer of the SA(X) unit to save memory costs.}.

Note that the three MCA layers above have not covered all the possible compositions. We have also explored other MCA variants like the symmetric architectures GA(X,Y)-GA(Y,X) and SGA(X,Y)-SGA(Y,X). However, these MCA variants do not report comparative performance, so we do not discuss them further due to space limitations.

\section{Modular Co-Attention Networks}
In this section, we describe the Modular Co-Attention Networks (MCAN) architecture for VQA.
We first explain the image and question feature representation from the input question and image. Then, we propose two deep co-attention models, namely \emph{stacking} and \emph{encoder-decoder}, which consists of multiple MCA layers cascaded in depth to gradually refine the attended image and question features. As we obtained the attended image and question features, we design a simple multimodal fusion model to fuse the multimodal features and finally feed them to a multi-label classifier to predict answer. An overview flowchart of MCAN is shown in Figure \ref{fig:mcan_flowchart}.

We name the MCAN model with the stacking strategy as MCAN$_{\mathrm{sk}}$-$L$ and the MCAN model with the encoder-decoder strategy as MCAN$_{\mathrm{ed}}$-$L$, where $L$ is the total number MCA layers cascaded in depth.

\subsection{Question and Image Representations}\label{sec:feature_rep}
The input image is represented as a set of regional visual features in a bottom-up manner \cite{anderson2017up-down}. These features are the intermediate features extracted from a Faster R-CNN model (with ResNet-101 as its backbone) \cite{ren2015faster} pre-trained on the Visual Genome dataset \cite{krishna2016visual}. We set a confidence threshold to the probabilities of detected objects and obtain a dynamic number of objects $m\in[10,100]$. For the $i$-th object, it is represented as a feature $x_i\in\mathbb{R}^{d_x}$ by mean-pooling the convolutional feature from its detected region. Finally, the image is represented as a feature matrix $X\in\mathbb{R}^{m \times d_x}$.

The input question is first tokenized into words, and trimmed to a maximum of 14 words similar to \cite{teney2017tips,kim2018bilinear}. Each word in the question is further transformed into a vector using the 300-D GloVe word embeddings \cite{pennington2014glove} pre-trained on a large-scale corpus. This results in a sequence of words of size $n \times$300, where $n\in [1,14]$ is the number of words in the question. The word embeddings are then passed through a one-layer LSTM network \cite{hochreiter1997long} with $d_y$ hidden units. In contrast to \cite{teney2017tips} which only uses the final state (\ie, the output feature for the last word) as the question feature, we maintain the output features for all words and output a question feature matrix $Y\in\mathbb{R}^{n \times d_y}$.

To deal with the variable number of objects $m$ and variable question length $n$, we use zero-padding to fill $X$ and $Y$ to their maximum sizes (\ie, $m=100$ and $n=14$, respectively). During training, we mask the padding logits with $-\infty$ to get zero probability before every softmax layer to avoid the underflow problem.

\subsection{Deep Co-Attention Learning}\label{sec:dcl}
Taking the aforementioned image features $X$ and the question features $Y$ as inputs, we perform deep co-attention learning by passing the input features though a deep co-attention model consisting of $L$ MCA layers cascaded in depth (denoted by MCA$^{(1)}$, MCA$^{(2)}$ ... MCA$^{(L)}$). Denoting the input features for MCA$^{(l)}$ as $X^{(l-1)}$ and $Y^{(l-1)}$ respectively, their output features are denoted by $X^{(l)}$ and $Y^{(l)}$, which are further fed to the MCA$^{(l+1)}$ as its inputs in a recursive manner.

\begin{equation}\label{eq:cau_cascade}
[X^{(l)},Y^{(l)}]= \mathrm{MCA}^{(l)}([X^{(l-1)},Y^{(l-1)}])
\end{equation}

For MCA$^{(1)}$, we set its input features $X^{(0)}=X$ and $Y^{(0)}=Y$, respectively.

\begin{figure}
    \centering
    \begin{subfigure}[h]{0.38\columnwidth}
        \includegraphics[width=\linewidth]{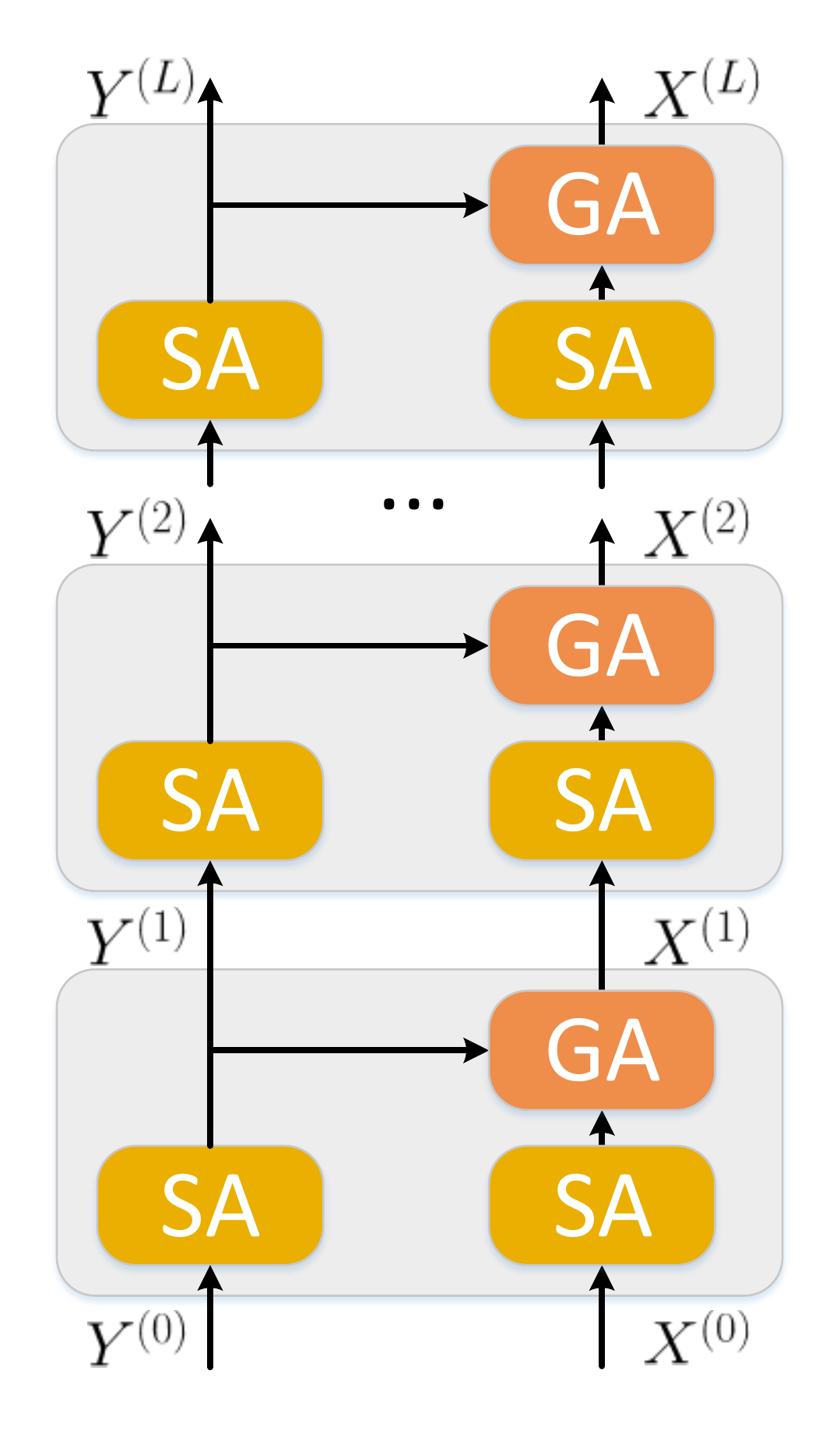}
        \caption{Stacking}\label{fig:stack}
    \end{subfigure}
    \begin{subfigure}[h]{0.38\columnwidth}
        \includegraphics[width=\linewidth]{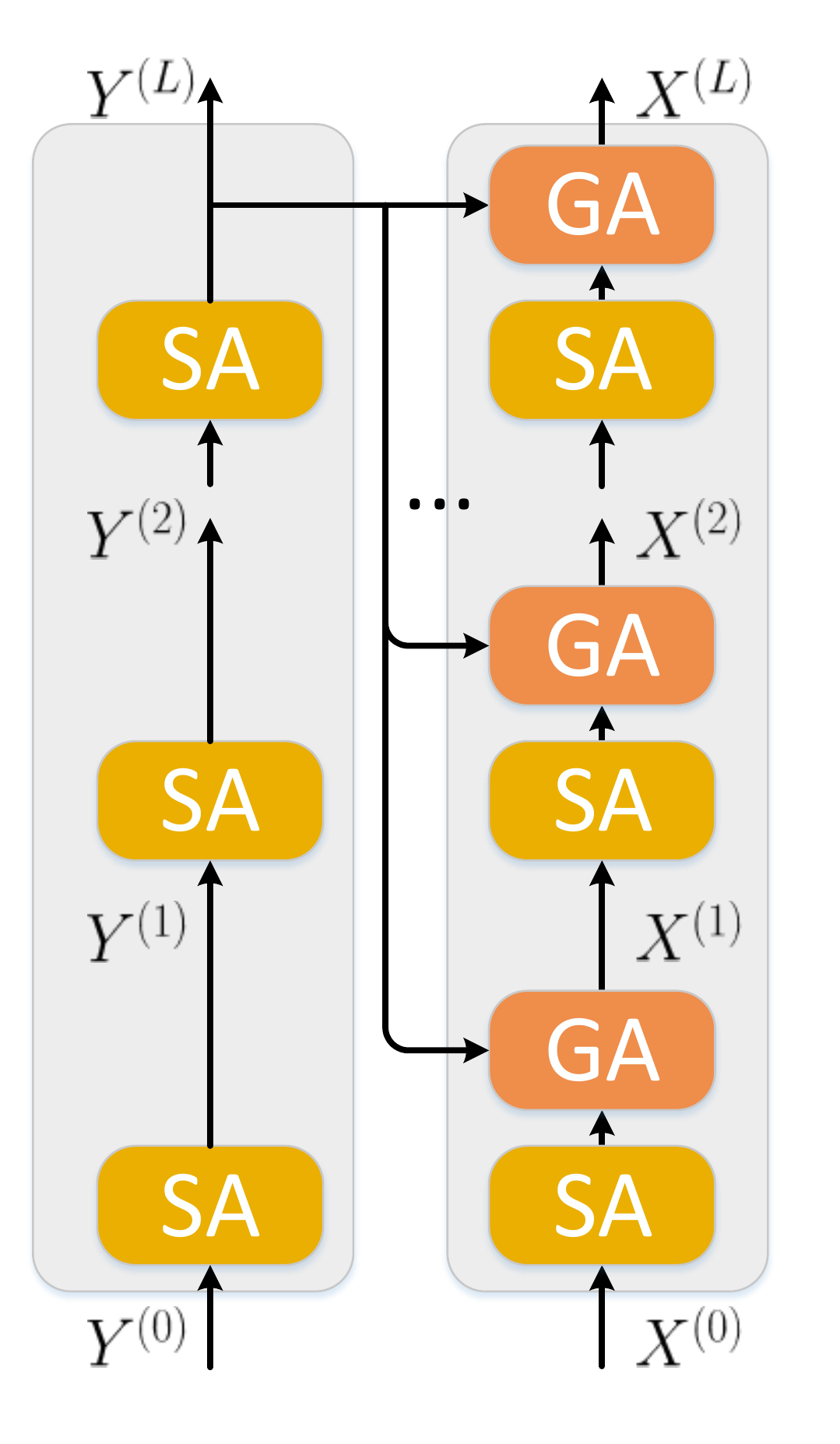}
        \caption{Encoder-Decoder}\label{fig:ed}
    \end{subfigure}
    \vspace{-5pt}
    \caption{Two deep co-attention models based on a cascade of MCA layers (\eg, SA(Y)-SGA(X,Y)).}
    \label{fig:deep_arch}
    \vspace{-10pt}
\end{figure}

Taking the SA(Y)-SGA(X,Y) layer as an example (the other two MCA layers proceed in the same manner), we formulate two deep co-attention models in Figure \ref{fig:deep_arch}.

The \emph{stacking} model (Figure \ref{fig:stack}) simply stacks $L$ MCA layers in depth and outputs $X^{(L)}$ and $Y^{(L)}$ as the final attended image and question features. The \emph{encoder-decoder} model (Figure \ref{fig:ed}) is inspired by the Transformer model proposed in \cite{vaswani2017attention}. It slightly modifies the stacking model by replacing the input features $Y^{(l)}$ of the GA unit in each MCA$^{(l)}$ with the question features $Y^{(L)}$ from the last MCA layer. The encoder-decoder strategy can be understood as an encoder to learn the attended question features $Y^{(L)}$ with $L$ stacked SA units and a decoder to use $Y^{(L)}$ to learn the attended image features $X^{(L)}$ with stacked SGA units.

The two deep models are of the same size with the same $L$. As a special case that $L=1$, the two models are strictly equivalent to each other.

\subsection{Multimodal Fusion and Output Classifier}
After the deep co-attention learning stage, the output image features $X^{(L)}=[x_1^{(L)};...;x_{m}^{(L)}]\in\mathbb{R}^{m \times d}$ and question features  $Y^{(L)}=[y_1^{(L)};...;y_{n}^{(L)}]\in\mathbb{R}^{n \times d}$ already contain rich information about the attention weights over the question words and image regions. Therefore, we design an attentional reduction model with a two-layer MLP (FC($d$)-ReLU-Dropout(0.1)-FC(1)) for $Y^{(L)}$ (or $X^{(L)}$) to obtain its attended feature $\tilde{y}$ (or $\tilde{x}$). Taking $X^{(L)}$ as an example, the attended feature $\tilde{x}$ is obtained as follows:
\begin{equation}\label{eq:self_att}
\begin{aligned}
\alpha &= \mathrm{softmax}(\mathrm{MLP}(X^{(L)})) \\
\tilde{x}&=\sum_{i=1}^m \alpha_ix_i^{(L)}
\end{aligned}
\end{equation}
where $\alpha=[\alpha_1,\alpha_2,...,\alpha_m]\in\mathbb{R}^m$ are the learned attention weights. We can obtain the attended feature $\tilde{y}$ for $Y^{(L)}$ using an independent attentional reduction model by analogy.

Using the computed $\tilde{y}$ and $\tilde{x}$, we design the linear multimodal fusion function as follows:
\begin{equation}\label{eq:fusion}
z = \mathrm{LayerNorm}(W_x^T\tilde{x} + W_y^T\tilde{y})
\end{equation}
where $W_x, W_y\in\mathbb{R}^{d \times d_z}$ are two linear projection matrices. $d_z$ is the common dimensionality of the fused feature. $\mathrm{LayerNorm}$ is used here to stabilize training \cite{ba2016layer}.


The fused feature $z$ is projected into a vector $s\in\mathbb{R}^N$ followed by a sigmoid function, where $N$ is the number of the most frequent answers in the training set. Following \cite{teney2017tips}, we use binary cross-entropy (BCE) as the loss function to train an $N$-way classifier on top of the fused feature $z$.

\captionsetup[subfigure]{font=small}

\captionsetup[subtable]{font=small}
\begin{table*}
    \scriptsize
	\begin{subtable}[t]{.34\textwidth}
		\centering
		\subcaption{\textbf{MCA Variants:} Accuracies of the MCAN model with different MCA variants under \textbf{one} layer. ID(Y)-GA(X,Y), SA(Y)-GA(X,Y) and SA(Y)-SGA(X,Y) denote the three MCA variants w/ or w/o the SA units for image and question (see Figure \ref{fig:cau_arch}). Since the stacking and the encoder-decoder strategies are equivalent under one layer, we do not distinguish them.}
        \begin{tabular}{l|cccc}
            \toprule
            Model & All & Y/N & Num & Other \\
            \midrule
            ID(Y)-GA(X,Y) & 64.8 & 82.5 & 44.7 & 56.7\\
            SA(Y)-GA(X,Y) & 65.2 & 82.9 & 44.8 & 57.1\\
            SA(Y)-SGA(X,Y) & \textbf{65.4} & \textbf{83.2} & \textbf{44.9} & \textbf{57.2} \\
            \bottomrule
        \end{tabular}
        \label{table:mcas}
    \end{subtable}
    \quad
    \scriptsize
    \begin{subtable}[t]{.26\textwidth}
		\centering
        \subcaption{\textbf{Stacking \textit{vs.} Encoder-decoder:} Overall accuracies and model sizes (\ie, number of parameters) of the MCAN$_{\mathrm{sk}}$-$L$ models and the MCAN$_{\mathrm{ed}}$-$L$ models, where number of layers $L\in\{2,4,6,8\}$. With the same $L$, the sizes of the two models are equal.}
		\begin{tabular}{l|cc|c}
            \toprule
            $L$ & MCAN$_{\mathrm{sk}}$ & MCAN$_{\mathrm{ed}}$ & Size  \\
            \midrule
            2 & 66.1& 66.2 & 27M\\
            4 & 66.7& 66.9 & 41M\\
            6 & 66.8& \textbf{67.2} & 56M\\
            8 & 66.8& \textbf{67.2} & 68M\\
            \bottomrule
        \end{tabular}
		\label{table:sk_ed}
	\end{subtable}
    \quad
    \scriptsize
	\begin{subtable}[t]{.36\textwidth}
		\centering
        \subcaption{\textbf{Question Representations:} Accuracies of the MCAN$_{\mathrm{ed}}$-6 model with different question representations. Rand$_{\mathrm{ft}}$ means the word embeddings are initialized randomly and then fine-tuned. PE denotes the positional encoding \cite{vaswani2017attention}. GloVe$_{\mathrm{pt+ft}}$ and GloVe$_{\mathrm{pt}}$ mean the word embeddings are pre-trained with GloVe, while GloVe$_{\mathrm{pt+ft}}$ is additionally fine-tuned.}
		\begin{tabular}{l|cccc}
            \toprule
            Model & All & Y/N & Num & Other \\
            \midrule
            Rand$_{\mathrm{ft}}$~+~PE & 65.6 & 83.0 & 47.9 & 57.1 \\
            GloVe$_{\mathrm{pt}}$~+~PE & 67.0 & 84.6 & \textbf{49.4} & 58.2\\
            GloVe$_{\mathrm{pt}}$~+~LSTM & 67.1 & \textbf{84.8} & \textbf{49.4} & 58.4\\
            GloVe$_{\mathrm{pt+ft}}$~+~LSTM & \textbf{67.2} & \textbf{84.8} & {49.3} & \textbf{58.6}\\
            \bottomrule
        \end{tabular}
		\label{table:ques_feat}
	\end{subtable}
	\caption{Ablation experiments for MCAN. All the reported results are evaluated on the \emph{val} split.}
    \label{table:aba}
\end{table*}

\captionsetup[subfigure]{font=small}
\begin{figure*}
    \centering
    \begin{subfigure}[h]{0.24\textwidth}
        \includegraphics[width=\linewidth]{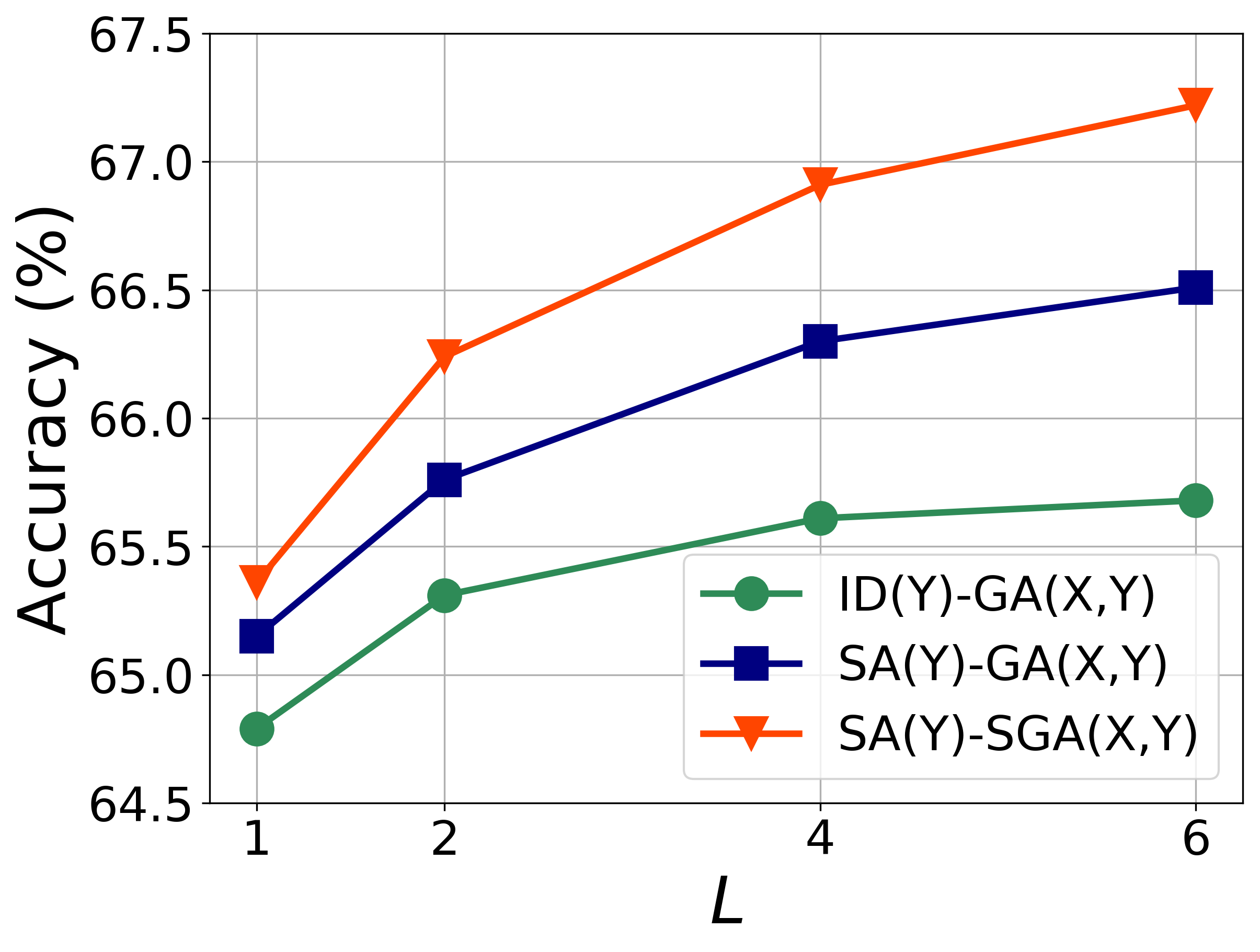}
        \caption{All}\label{fig:acc_mcus_overall}
    \end{subfigure}
    \begin{subfigure}[h]{0.24\textwidth}
        \includegraphics[width=\linewidth]{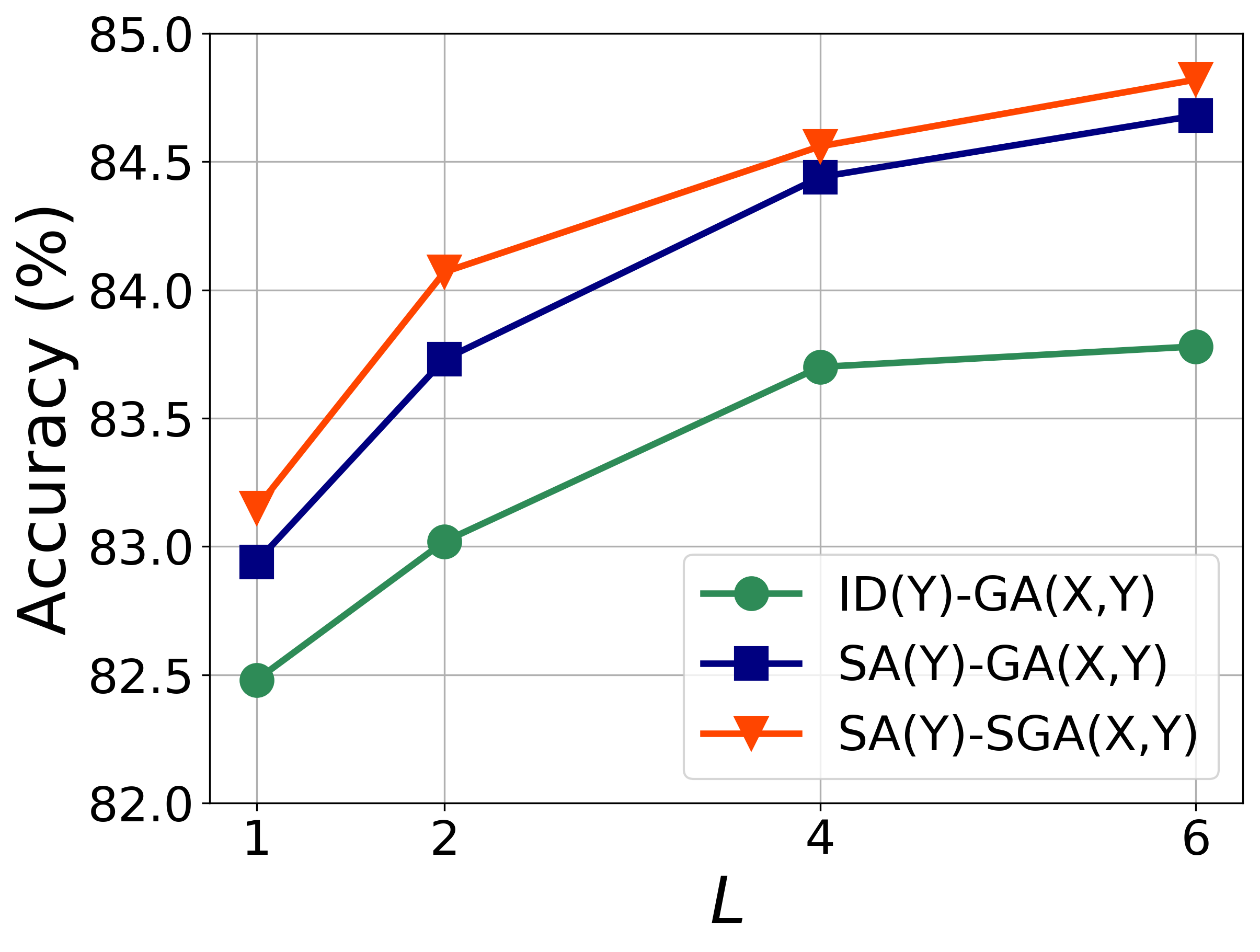}
        \caption{Y/N}\label{fig:acc_mcus_yn}
    \end{subfigure}
    \begin{subfigure}[h]{0.24\textwidth}
            \includegraphics[width=\linewidth]{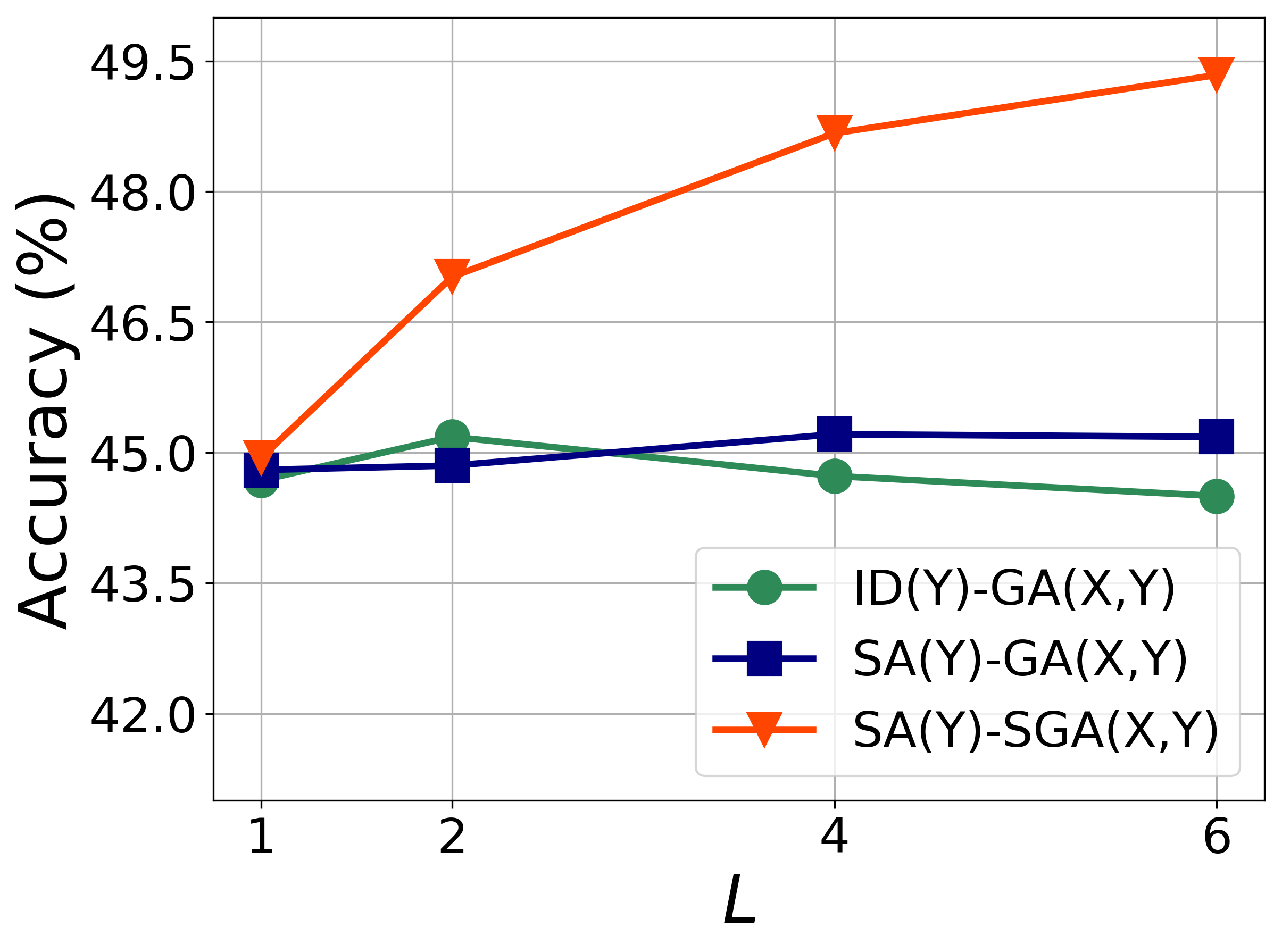}
        \caption{Num}\label{fig:acc_mcus_number}
    \end{subfigure}
    \begin{subfigure}[h]{0.24\textwidth}
        \includegraphics[width=\linewidth]{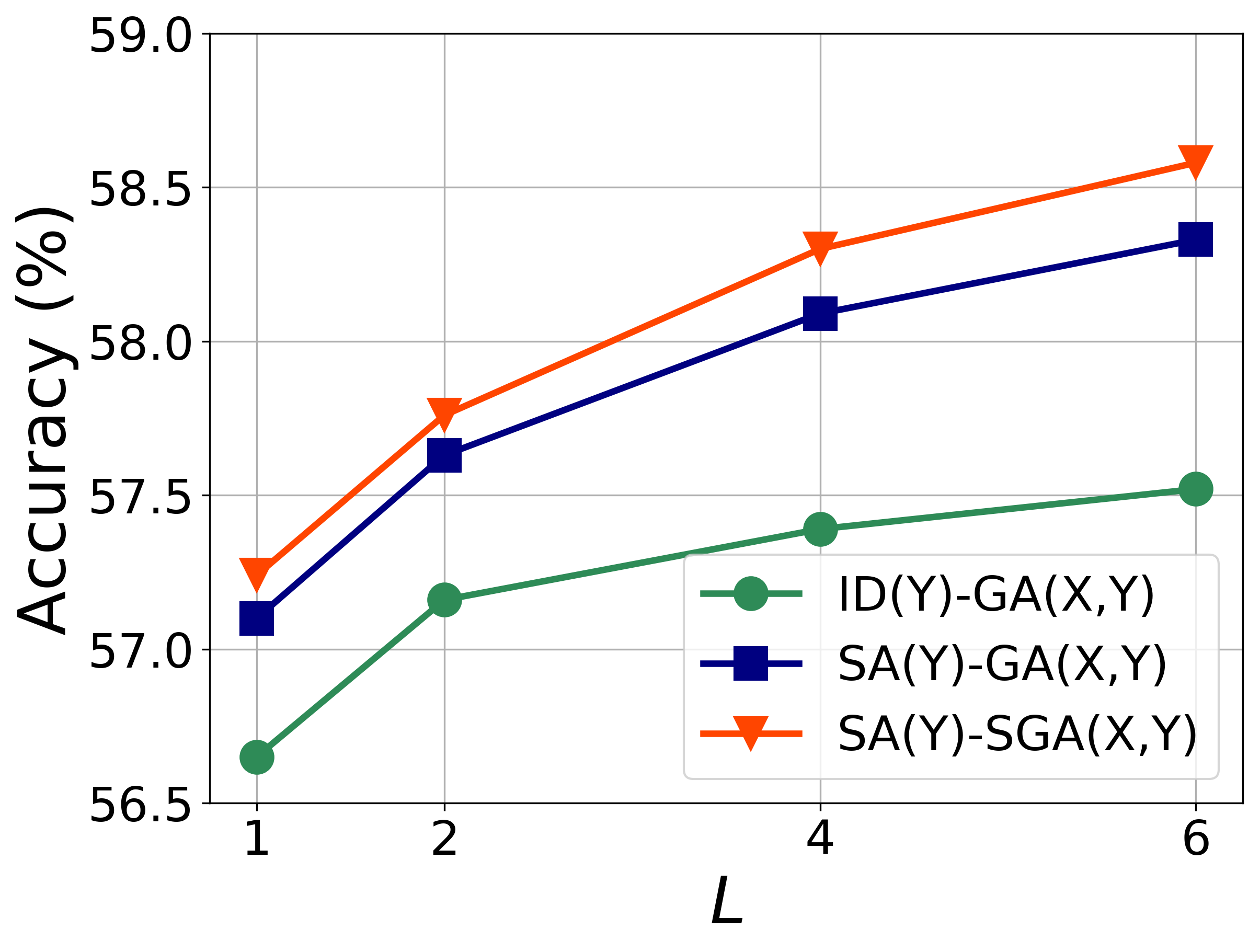}
        \caption{Other}\label{fig:acc_mcus_other}
    \end{subfigure}
    \caption{The overall and per-type accuracies of the MCAN$_{\mathrm{ed}}$-$L$ models equipped with different MCA variants, where the number of layers $L\in\{1,2,4,6\}$. All the reported results are evaluated on the \emph{val} split.}
    \vspace{-10pt}
    \label{fig:acc_mcus}
\end{figure*}

\section{Experiments}
In this section, we conduct experiments to evaluate the performance of our MCAN models on the largest VQA benchmark dataset, VQA-v2 \cite{goyal2016making}. Since the different MCA variants and deep co-attention models may influence final performance, we perform extensive quantitative and qualitative ablation studies to explore the reasons why MCAN performs well. Finally, with the optimal hyper-parameters, we compare our best model with current state-of-the-art models under the same settings.

\subsection{Datasets}
\textbf{\textit{VQA-v2}} is the most commonly used VQA benchmark dataset \cite{goyal2016making}. It contains human-annotated question-answer pairs relating to the images from the MS-COCO dataset \cite{lin2014microsoft}, with 3 questions per image and 10 answers per question. The dataset is split into three: \emph{train} (80k images and 444k QA pairs); \emph{val} (40k images and 214k QA pairs); and \emph{test} (80k images and 448k QA pairs). Additionally, there are two test subsets called \emph{test-dev} and \emph{test-standard} to evaluate model performance online. The results consist of three per-type accuracies (\emph{Yes/No}, \emph{Number}, and \emph{Other}) and an overall accuracy.

\subsection{Implementation Details}
The hyper-parameters of our model used in the experiments are as follows. The dimensionality of input image features $d_x$, input question features $d_y$, and fused multimodal features $d_z$ are 2,048, 512, and 1,024, respectively. Following the suggestions in \cite{vaswani2017attention}, the latent dimensionality $d$ in the multi-head attention is 512, the number of heads $h$ is set to 8, and the latent dimensionality for each head is $d_h=d/h=64$. The size of the answer vocabulary is set to $N=3,129$ using the strategy in \cite{teney2017tips}. The number of MCA layers is $L\in\{1,2,4,6,8\}$.

To train the MCAN model, we use the Adam solver \cite{kingma2014adam} with $\beta_1=0.9$ and $\beta_2=0.98$. The base learning rate is set to $\mathrm{min}(2.5te^{-5},1e^{-4})$, where $t$ is the current epoch number starting from 1. After 10 epochs, the learning rate is decayed by 1/5 every 2 epochs. All the models are trained up to 13 epochs with the same batch size 64. For the results on the \emph{val} split, only the \emph{train} split is used for training. For the results on the \emph{test-dev} or \emph{test-standard} splits, both \emph{train} and \emph{val} splits are used for training, and a subset of VQA samples from Visual Genome \cite{krishna2016visual} is also used as the augmented dataset to facilitate training.

\captionsetup[subfigure]{font=small}
\begin{figure*}
    \begin{subfigure}[h]{0.23\textwidth} 
        \includegraphics[width=\linewidth]{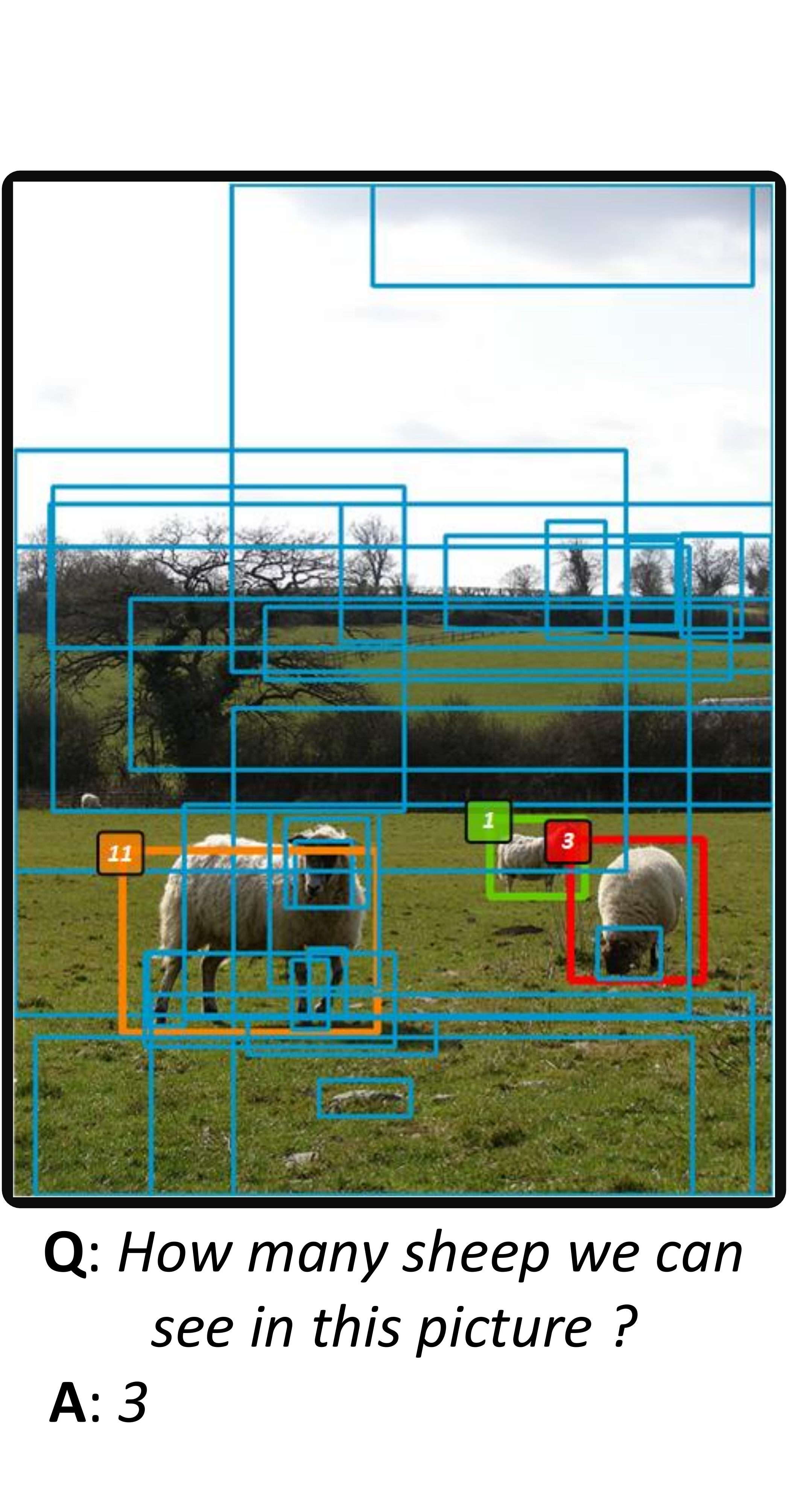}
        \label{fig:vis1}
    \end{subfigure}
    \begin{subfigure}[h]{0.335\textwidth} 
        \includegraphics[width=\linewidth]{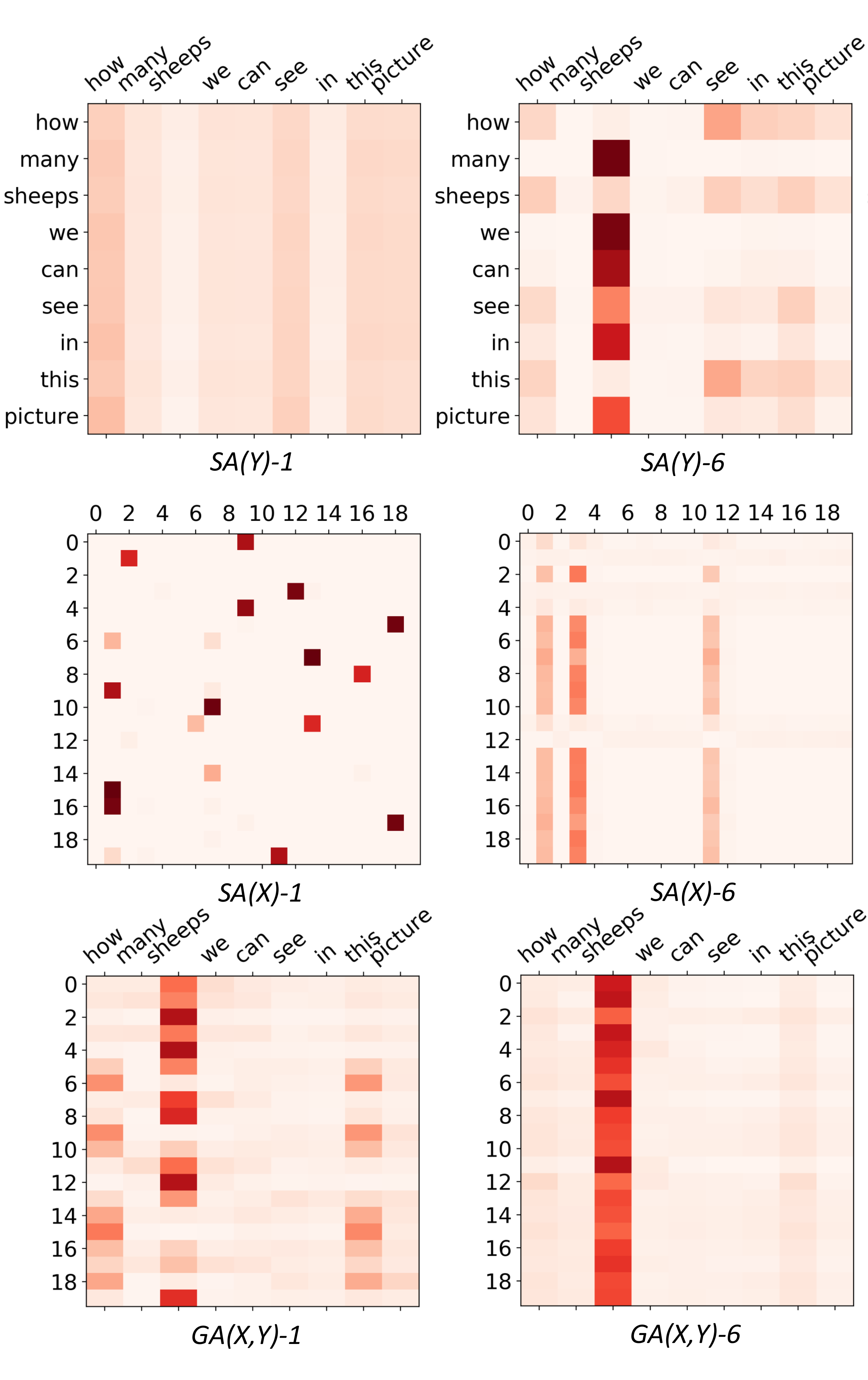}
        \caption{Encoder-Decoder~~~(\textbf{P}: 3)}\label{fig:vis2}
    \end{subfigure}
    \quad
    \begin{subfigure}[h]{0.38\textwidth} 
        \includegraphics[width=\linewidth]{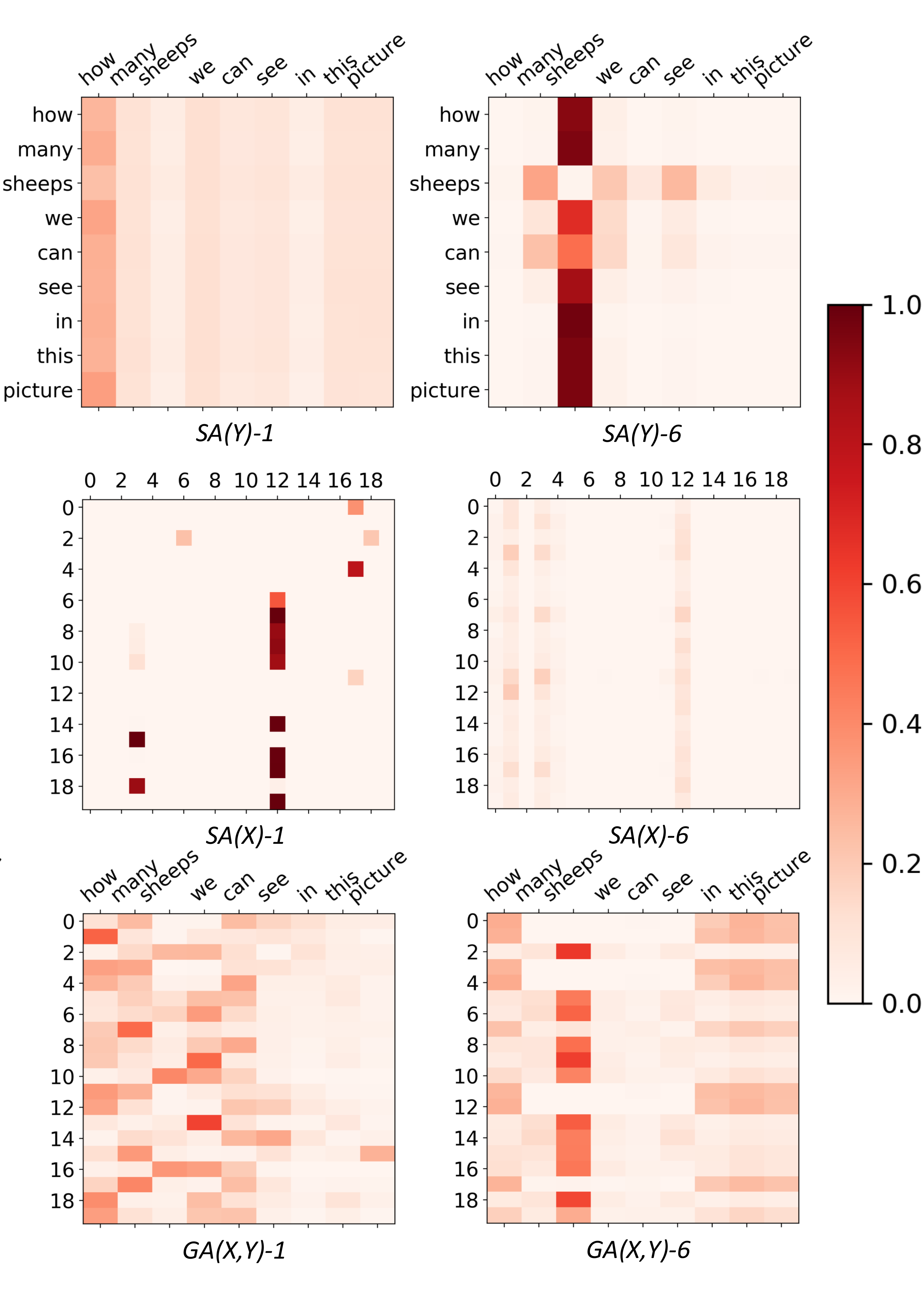}
        \caption{Stacking~~~(\textbf{P}: 3)}\label{fig:vis3}
    \end{subfigure}
    \caption{Visualizations of the learned attention maps ($\mathrm{softmax}(qK/\sqrt{d})$) of the attention units from typical layers. SA(Y)-$l$, SA(X)-$l$ and GA(X,Y)-$l$ denote the question self-attention, image self-attention, and image guided-attention from the $l$-th layer, respectively. {Q}, {A}, {P} denote the question, answer and prediction respectively. The index within [0-19] shown on the axes of the attention maps corresponds to each object in the image (20 objects in total) . For better visualization effect, we highlight three objects in the image that are related to the answer (\ie, sheep).}
    \label{fig:vis}
    \vspace{-8pt}
\end{figure*}

\subsection{Ablation Studies}\label{sec:ablation}
We run a number of ablations to investigate the reasons why MCAN is effective. The results shown in Table \ref{table:aba} and Figure \ref{fig:acc_mcus} are discussed in detail below.
\\
\textbf{MCA Variants:} From the results in Table \ref{table:mcas}, we can see that SA(Y)-GA(X,Y) outperforms ID(Y)-GA(X,Y) for all answer types. This verifies that modeling self-attention for question features benefits VQA performance, which is consistent with previous works \cite{yu2018beyond}. Moreover, we can see that SA(Y)-SGA(X,Y) also outperforms SA(Y)-GA(X,Y). This reveals, for the first time, that modeling self-attention for image features is meaningful. Therefore, we use SA(Y)-SGA(X,Y) as our default MCA in the following experiments unless otherwise stated.
\\
\textbf{Stacking \textit{vs.} Encoder-Decoder:} From the results in Table \ref{table:sk_ed}, we can see that with increasing $L$, the performances of both deep co-attention models steadily improve and finally saturate at $L=6$. The saturation can be explained by the unstable gradients during training when $L> 6$, which makes the optimization difficult. Similar observations are also reported by \cite{bapna2018training}. Furthermore, the encoder-decoder model steadily outperforms the stacking model, especially when $L$ is large. This is because the learned self-attention from an early SA(Y) unit is inaccurate compared to that from the last SA(Y) unit. Directly feeding it to a GA(X,Y) unit may damage the learned guided-attention for images. The visualization in $\S$\ref{sec:vis} supports this explanation. Finally, MCAN is much more parametric-efficient than other approaches, with MCAN$_{\mathrm{ed}}$-2 (27M) reporting a 66.2$\%$ accuracy, BAN-4 (45M) a 65.8$\%$ accuracy \cite{kim2018bilinear}, and MFH (116M) a 65.7$\%$ accuracy \cite{yu2018beyond}. More in-depth comparisons can be found in the supplementary material.
\\
\textbf{MCA \textit{vs.} Depth:} In Figure \ref{fig:acc_mcus}, we show the detailed performance of MCAN$_{\mathrm{ed}}$-$L$ with different MCA variants. With increasing $L$, the performance gaps between the three variants increases. Furthermore, an interesting phenomenon occurs in Figure \ref{fig:acc_mcus_number}. When $L=6$, the \emph{number} type accuracy of the ID(Y)-GA(X,Y) and SA(Y)-GA(X,Y) models are nearly identical, while the SA(Y)-SGA(X,Y) model reports a 4.5-point improvement over them. This verifies that self-attention for images plays a key role in object counting.
\\
\textbf{Question Representations:} Table \ref{table:ques_feat} summarizes ablation experiments on different question representations. We can see that using the word embeddings pre-trained by GloVe \cite{pennington2014glove} significantly outperforms that by random initialization. Other trick like fine-tuning the GloVe embeddings or replacing the position encoding \cite{vaswani2017attention} with a LSTM network to model the temporal information can slightly improve the performance further.

\begin{figure*}
\begin{center}
\includegraphics[width=0.99\textwidth]{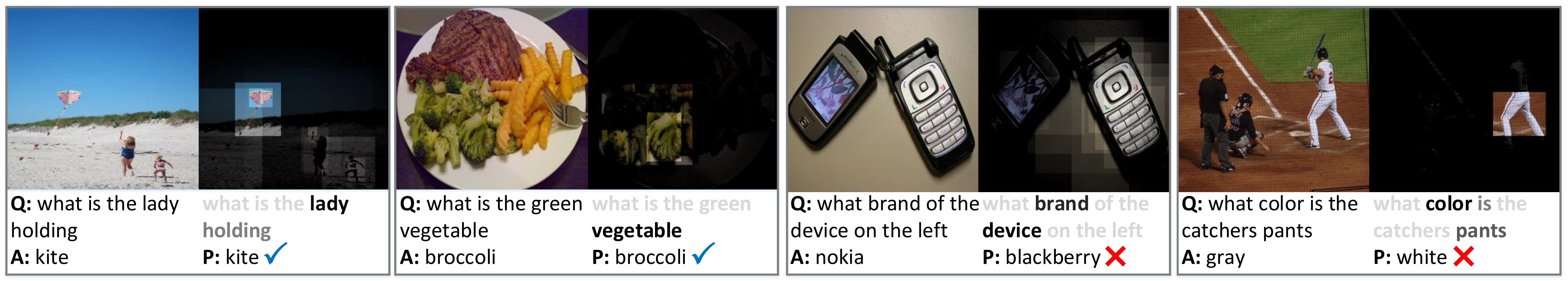}
\vspace{-5pt}
\caption{Typical examples of the learned image and question attentions by Eq.(\ref{eq:self_att}). For each example, the image, question (Q) and answer (A) are presented on the left; the learned image attention, question attention and prediction (P) are presented next to them. The brightness of regions and darkness of words represent their importance in the attention weights.}
\vspace{-10pt}
\label{fig:vis_coatt}
\end{center}
\end{figure*}

\subsection{Qualitative Analysis}\label{sec:vis}
In Figure \ref{fig:vis}, we visualize the learned attentions from MCAN$_{\mathrm{sk}}$-6 and MCAN$_{\mathrm{ed}}$-6. Due to space limitations, we only show one example and visualize six attention maps from different attention units and different layers. More visualizations can be found in the supplementary material. From the results, we have the following observations.
\\
\textbf{Question Self-Attention SA(Y):} The attention maps of SA(Y)-1 form vertical stripes, and the words like `how' and `see' obtain large attention weights. This unit acts as a question type classifier. Besides, the large values in the attention maps of SA(Y)-6 occur in the column `sheep'. This reveals that all the attended features tend to use the feature of `sheep' for reconstruction. That is to say, the keyword `sheep' is identified correctly.
\\
\textbf{Image Self-Attention SA(X):} Values in the attention maps of SA(X)-1 are uniformly distributed, suggesting that the key objects for sheep are unclear. The large values in the attention maps of SA(X)-6 occur on the 1st, 3rd, and 11th columns, which correspond to the three sheep in the image. This explains why introducing SA(X) can greatly improve object counting performance.
\\
\textbf{Question Guided-Attention GA(X,Y):} The attention maps of GA(X,Y)-1 do not focus on the current objects in the image; and the attention maps of GA(X,Y)-6 tend to focus on all values in the `sheep' column. This can be explained by the fact that the input features have been reconstructed by the sheep features in SA(X)-6. Moreover, the GA(X,Y) units of the stacking model contain much more noise than the encoder-decoder model. This verifies our hypothesis presented in $\S$\ref{sec:ablation}.

In Figure \ref{fig:vis_coatt}, we also visualize the final image and question attentions learned by Eq.(\ref{eq:self_att}). For the correctly predicted examples, the learned question and image attentions are usually closely focus on the key words and the most relevant image regions (\eg, the word `holding' and the region of `hand' in the first example, and the word `vegetable' and the region of `broccoli' in the second example). From the incorrect examples, we can draw some weaknesses of our approach. For example, it occasionally makes mistakes in distinguishing the key words in questions (\eg, the word `left' in the third example and the word `catcher' in the last example). These observations are useful to guide further improvements in the future.

\begin{table}
\centering
\footnotesize
\caption{Accuracies of \textbf{single-model} on the \emph{test-dev} and \emph{test-standard} splits to compare with the state-of-the-art methods. All the methods use the same bottom-up attention visual features \cite{anderson2017up-down}, and are trained on the \emph{train+val+vg} sets (\emph{vg} denotes the augmented VQA samples from Visual Genome). The best results on both splits are bolded.}
\label{table:sota}
\begin{tabular}{l|cccc|c}
\toprule
\multirow{2}{*}{Model}& \multicolumn{4}{c|}{Test-dev} &Test-std \\
\cmidrule{2-6}
 & All & Y/N & Num & Other & All  \\
\midrule
Bottom-Up \cite{teney2017tips} & 65.32& 81.82& 44.21& 56.05 &65.67\\
MFH \cite{yu2018beyond} & 68.76 &84.27 &49.56 &59.89 & -\\
BAN \cite{kim2018bilinear} &69.52 &85.31& 50.93&  60.26&-\\
BAN+Counter \cite{kim2018bilinear}& 70.04 & 85.42& \textbf{54.04}&60.52&70.35 \\
\midrule
MCAN$_{\mathrm{ed}}$-6 & \textbf{70.63} & \textbf{86.82} & 53.26 & \textbf{60.72}  &\textbf{70.90}  \\
\bottomrule
\end{tabular}
\vspace{-10pt}
\end{table}
\subsection{Comparison with State-of-the-Art}
By taking the ablation results into account, we compare our best single model MCAN$_{\mathrm{ed}}$-6 with the current state-of-the-art methods in Table \ref{table:sota}. Using the same bottom-up attention visual features \cite{anderson2017up-down}, MCAN$_{\mathrm{ed}}$-6 significantly outperforms the current best approach BAN \cite{kim2018bilinear} by 1.1 points in terms of overall accuracy. Compared to BAN+Counter \cite{kim2018bilinear}, which additionally introduces the counting module \cite{zhang2018learning} to significantly improve object counting performance, our model is still 0.6 points higher. Moreover, our method obtains comparable object counting performance (\ie, the number type) to BAN+Counter, and in doing so does not use any auxiliary information like the bounding-box coordinates of each object \cite{zhang2018learning}. This suggests that MCAN is more general that can naturally learn to \emph{deduplicate} the redundant objects based on the visual features alone. The comparative results with model ensembling are demonstrated in the supplementary material.

\section{Conclusions}
In this paper, we present a novel deep Modular Co-Attention Network (MCAN) for VQA. MCAN consists of a cascade of modular co-attention (MCA) layers, each of which consists of the self-attention and guided-attention units to model the intra- and inter-modal interactions synergistically. By stacking MCA layers in depth using the encoder-decoder strategy, we obtain a deep MCAN model that achieves new state-of-the-art performance for VQA.

\section*{Acknowledgments}
This work was supported in part by National Natural Science Foundation
of China under Grant 61702143, Grant 61836002, Grant 61622205, and in part by the Australian Research Council Projects under Grant FL-170100117, Grant DP-180103424 and Grant  IH-180100002.

{\small
\bibliographystyle{ieee_fullname}
\bibliography{mcan}
}

\section*{Appendix}

\appendix
\section{Model Ensembling}
To compare MCAN to the best results on VQA-v2 leaderboard\footnote{\url{https://visualqa.org/roe.html}}, we train 4 MCAN$_{\mathrm{ed}}$-6 models with slightly different hyper-parameters for ensemble. The comparative results in Table \ref{table:sota} indicate that MCAN surpasses the top most solutions on the leaderboard. It is worth noting that our solution only use the basic bottom-up attention visual features \cite{anderson2017up-down} and much fewer models for ensemble.

\section{Comparisons of Model Stability and Computational Costs}
We compare MCAN$_{\mathrm{ed}}$-6 with the best two approaches (MFH \cite{yu2018beyond} and BAN-8 \cite{kim2018bilinear}) in Table \ref{table:1} in terms of overall accuracy $\pm$std, number of parameters and FLOPs, respectively. The accuracies are reported on the \emph{val} split, and the standard deviation for each method is calculated by training three models with the same architecture but different initializations. The FLOPs are calculated for one testing sample. We can see that MCAN$_{\mathrm{ed}}$-6 outperforms the counterparts in both accuracy and stability, and is more parameteric- and computational-efficient at the same time.

\section{More Visualized Results}
\begin{figure*}
\begin{center}
\includegraphics[width=0.99\textwidth]{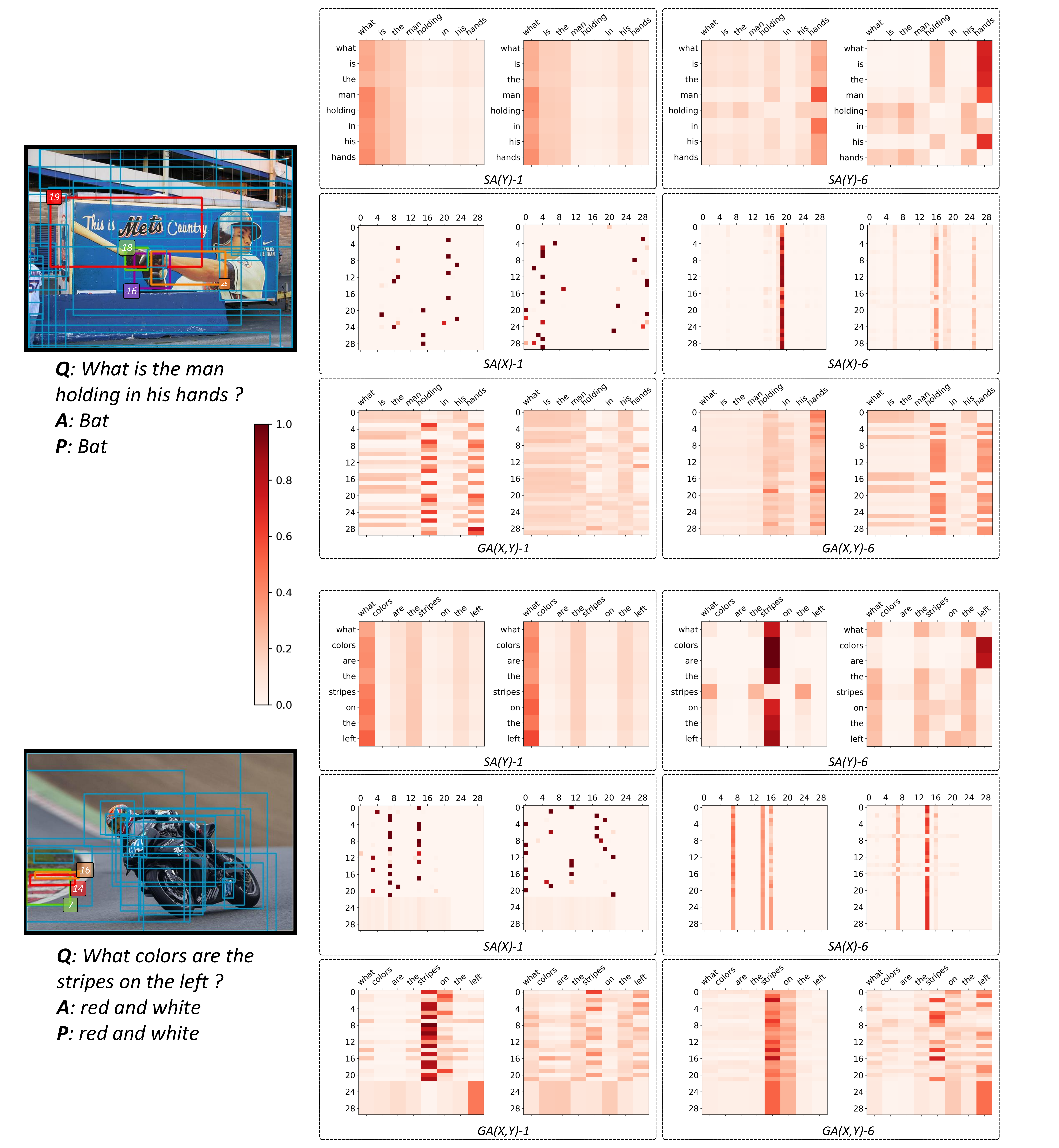}
\vspace{-5pt}
\caption{Two examples of the learned attention maps from typical attention units and layers. For each attention unit (within the box), we show two attention maps from different heads.}
\vspace{-15pt}
\label{fig:amap2}
\end{center}
\end{figure*}

Similar to Figure \ref{fig:vis} in the main text, we visualize the learned attentions of two more examples from MCAN$_{\mathrm{ed}}$-6 in Figure \ref{fig:amap2}. For each example, we visualize the attention maps from three attention units (SA(X), SA(Y), GA(X,Y)) and from two layers (1st and 6th). For each unit, we show the attention maps from 2 parallel heads (8 heads in total). From the results, we have the similar observations and explanations to those in the main text. The visualized attentions can well explain the reasoning process of MCAN to predict the correct answers. Furthermore, we find that different heads may provide complementary information to benefit VQA performance, which is similar to the `multi-glimpses' strategy in existing VQA approaches \cite{fukui2016multimodal,yu2018beyond}.

\begin{table}
\centering
\footnotesize
\caption{Accuracies of \textbf{model ensembling} on the \emph{test-standard} split to compare with the best solutions in VQA-Challenge 2018. $R$ denotes the rank of the corresponding team. \# denotes the number of used models for ensembling.}
\label{table:sota}
\begin{tabular}{c|l|c|cccc}
\toprule
$R$& Team Name &\# & All & Y/N & Num & Other\\
\midrule
5& MIL-UT & -& 71.16 & 87.00 & 52.6 & 61.62  \\
4& CASIA-IVA & -& 71.31 & 86.98 & 51.05 & 62.31  \\
3& SNU-BI & 15& 71.84 &	 87.22 & 54.37 & 62.45  \\
2& HDU-UCAS-USYD & 12 & 72.09 & 87.61 & 51.92 & 63.19  \\
1& FAIR A-STAR &30  & 72.25 & 87.82 & 51.59 &\textbf{63.43} \\
\midrule
& MCAN (Ours) & 4 &\textbf{72.45} &\textbf{88.29} & \textbf{54.38} & {62.80}  \\
\bottomrule
\end{tabular}
\end{table}

\begin{table}
\centering
\small
\caption{Comparison of model stability and computational costs to the state-of-the-art on \emph{val} split of VQA-v2.}
\label{table:1}
\begin{tabular}{c|c|c|c}
\toprule
 & MFH \cite{yu2018beyond}& BAN-8 \cite{kim2018bilinear}& MCAN$_{\mathrm{ed}}$-6 \\
 \midrule
Acc. $\pm$ std. (\%) &65.65$\pm$0.05 &66.04$\pm$0.08&67.23$\pm$0.01\\
\#Params ($\times 10^6$) & 116 & 79 & 56\\
FLOPs ($\times 10^9$)  & 4.4 &3.3& 2.8 \\
\bottomrule
\end{tabular}
\end{table}

\end{document}